%% file: main.tex
\newif\ifisarxiv
\isarxivtrue

\documentclass{article}

\usepackage{microtype}
\usepackage{graphicx}
\usepackage{subfigure}
\usepackage{booktabs} 
\usepackage{adjustbox}
\usepackage{multirow}
\usepackage{enumitem}
\usepackage{tikz}
\usepackage{xfrac}
\usepackage[compact]{titlesec}

\usetikzlibrary{shapes.geometric, arrows}
\usetikzlibrary{decorations.pathreplacing, chains}
\usetikzlibrary{fit}
\usetikzlibrary{calc}
\usetikzlibrary{backgrounds}
\usetikzlibrary{patterns}
\usetikzlibrary{positioning,arrows.meta,bending}

\usepackage{hyperref}
\usepackage{listings}

\ifisarxiv
\usepackage{fullpage}
\newcommand{\citet}[1]{\cite{#1}}
\renewcommand{\vspace}[1]{}
\newlength{\mywidth}
\RequirePackage{algorithm}
\RequirePackage{algorithmic}
\else
\usepackage[accepted]{icml2022}
\usepackage{icml2022}
\newlength{\mywidth}
\fi

\input{common}

\newcommand{\method}{GACT\xspace}
\newcommand{\code}[1]{\tt{#1}}

\ifisarxiv
\else
\icmltitlerunning{\method:  Activation Compressed Training for Generic Network Architectures}
\fi

\ifisarxiv
\title{\method:  Activation Compressed Training for Generic Network Architectures}
\author{%
  Xiaoxuan Liu, Lianmin Zheng, Dequan Wang, Yukuo Cen, \\
  Weize Chen, Xu Han, Jianfei Chen, Zhiyuan Liu, Jie Tang, \\
  Joey Gonzalez, Michael Mahoney, Alvin Cheung \\
  University of California, Berkeley\\
  \texttt{xiaoxuan\_liu@berkeley.edu, jianfeic@tsinghua.edu.cn} \\
}
\date{}
\fi

\usepackage{enumitem}
\begin{document}

\setlength{\textfloatsep}{0.1cm}
\setlength{\floatsep}{0.1cm}
\setlength{\intextsep}{0pt}
\setlength{\abovedisplayskip}{1pt}
\setlength{\belowdisplayskip}{1pt}
\setlist{nolistsep}

\ifisarxiv
\maketitle
\else
\twocolumn[
\icmltitle{\method:  Activation Compressed Training for Generic Network Architectures
}




\begin{icmlauthorlist}
\icmlauthor{Xiaoxuan Liu}{ucb}
\icmlauthor{Lianmin Zheng}{ucb}
\icmlauthor{Dequan Wang}{ucb}
\icmlauthor{Yukuo Cen}{thu}
\icmlauthor{Weize Chen}{thu}
\icmlauthor{Xu Han}{thu}
\icmlauthor{Jianfei Chen}{thu}
\icmlauthor{Zhiyuan Liu}{thu}
\icmlauthor{Jie Tang}{thu}
\icmlauthor{Joseph E. Gonzalez}{ucb}
\icmlauthor{Michael W. Mahoney}{ucb,icsi,lbnl}
\icmlauthor{Alvin Cheung}{ucb}
\end{icmlauthorlist}

\icmlaffiliation{ucb}{UC Berkeley}
\icmlaffiliation{thu}{Dept. of Comp. Sci. \& Tech., Institute for AI, Tsinghua-Bosch Joint Center for ML, BNRist Center, State Key Lab for Intell. Tech. \& Sys., Tsinghua University}
\icmlaffiliation{icsi}{ICSI}
\icmlaffiliation{lbnl}{LBNL}

\icmlcorrespondingauthor{Jianfei Chen}{jianfeic@tsinghua.edu.cn}

\icmlkeywords{Machine Learning, ICML}

\vskip 0.3in
]



\printAffiliationsAndNotice{}  
\fi

\begin{abstract}
Training large neural network (NN) models requires extensive memory resources, and Activation Compressed Training (ACT) is a promising approach to reduce training memory footprint. 
This paper presents \method, an ACT framework to support a broad range of machine learning tasks for generic NN architectures with limited domain knowledge.
By analyzing a linearized version of ACT's approximate gradient, we prove the convergence of \method without prior knowledge on operator type or model architecture. 
To make training stable, 
we propose an algorithm that decides the compression ratio for each tensor by estimating its impact on the gradient at run time.
We implement \method as a PyTorch library that readily applies to any NN architecture.
\method reduces the activation memory for convolutional NNs, transformers, and graph NNs by up to 8.1$\times$, enabling training with a 4.2$\times$ to 24.7$\times$ larger batch size, with negligible accuracy loss. We implement \method as a PyTorch library at \url{https://github.com/LiuXiaoxuanPKU/GACT-ICML}.


\end{abstract}

\input{theorems}
\input{1-introduction}
\input{2-related}
\input{3-theory}

\input{4-adaptive}
\input{5-system}

\input{6-experiments}

\section{Conclusion}
This paper presents \method, an ACT framework for generic NN architectures. We prove the convergence of \method without prior knowledge about operator type or network architecture by analyzing a linearized approximation of ATC's gradients. With the adaptive algorithm, \method achieves negligible accuracy loss on various tasks, reducing activation memory by up to 8.1$\times$ and enabling training with up to 24.7$\times$ batch size compared with full precision training.

\section*{Acknowledgements}
This work was supported by the National Key Research and Development Project of China (No. 2021ZD0110502); NSF of China Project (No. 62106120), by the National Science Foundation through grants IIS-1955488, IIS-2027575, CCF-1723352, ARO W911NF2110339, ONR N00014-21-1-2724, and DOE award DE-SC0016260.
We would also like to acknowledge partial support from DARPA, IARPA, the Sloan Foundation, NSF, and ONR.
Our conclusions do not necessarily reflect the position or the policy of our sponsors, and no official endorsement should be inferred.

\bibliography{refs}

\ifisarxiv
\bibliographystyle{unsrt}
\else
\bibliographystyle{icml2022}
\fi

\newpage
\appendix
\onecolumn
\input{aa-theorems}

\input{experiment-setup}
\input{experiment-baseline-accuracy}

\end{document}

%% file: common.tex
\usepackage{color}
\usepackage{bm}
\usepackage{hyperref}
\usepackage{amsmath,amsfonts,amsthm}
\usepackage{blindtext}
\usepackage{listings}
\usepackage{array}

\usepackage{ifthen}

\newif\ifdebug
\debugfalse

\ifdebug
\usepackage[textsize=tiny,color=lightgray,textwidth=2cm]{todonotes}
\paperwidth=\dimexpr \paperwidth + 4cm\relax
\oddsidemargin=\dimexpr\oddsidemargin + 2cm\relax
\evensidemargin=\dimexpr\evensidemargin + 2cm\relax
\else
\usepackage[disable]{todonotes}
\fi

\newtheorem{theorem}{Theorem}

\newtheorem{proposition}{Proposition}

\newcommand{\ceil}[1]{\left\lceil #1 \right\rceil}
\newcommand{\floor}[1]{\left\lfloor #1 \right\rfloor}

\newcommand{\argmin}{\operatornamewithlimits{argmin}}

\newcommand{\E}[1]{\mathbb{E}\left[#1\right]}
\newcommand{\Es}[2]{\mathbb{E}_{#1}\left[#2\right]}
\newcommand{\Econd}[2]{\mathbb{E}\left[#1\;\middle|\;#2\right]}
\newcommand{\Var}[1]{\mathrm{Var}\left[#1\right]}
\newcommand{\Vars}[2]{\mathrm{Var}_{#1}\left[#2\right]}
\newcommand{\Cov}[2]{\mathrm{Cov}\left[#1,#2\right]}

\newcommand{\Varcond}[2]{\mathrm{Var}\left[#1\;\middle|\;#2\right]}

\newcommand{\Lc}{\mathcal L}

\newcommand{\Xc}{\mathcal X}

\newcommand{\Eb}{\mathbb E}

\newcommand{\Ib}{\mathbb I}

\newcommand{\Rb}{\mathbb R}

\newcommand{\norm}[1]{\left\lVert#1\right\rVert}

\newcommand{\iprod}[1]{\langle#1\rangle}

\makeatletter
\newtheorem*{rep@theorem}{\rep@title}
\newcommand{\newreptheorem}[2]{%
	\newenvironment{rep#1}[1]{%
		\def\rep@title{#2 \ref{##1}}%
		\begin{rep@theorem}}%
		{\end{rep@theorem}}}
\makeatother

%% file: theorems.tex
\newcommand{\propa}{

}

\newcommand{\propb}{\label{prop:var-order}

}

%% file: 1-introduction.tex
\section{Introduction}
In recent years, we have witnessed the trend of using larger and larger neural network (NN) models to deliver improved accuracy and generalization
in various machine learning tasks~\cite{devlin2018bert, fedus2021switch}. 
However, training these models requires a considerable amount of on-device GPU memory. 
Unfortunately, the increase of GPU memory capacity has been relatively slow, leading to a fundamental barrier to the development of large NN models.

Activation Compressed Training (ACT) is a promising approach to reduce the memory footprint of models during training. As all layers' activations need to be kept in the memory for computing the gradients during training,
ACT reduces memory consumption by compressing these saved activations. 
Prior work~\cite{chakrabarti2019backprop, fu2020don, chen2021actnn, evans2021ac} has shown the effectiveness of ACT by reducing activation footprint by up to $12\times$ with 2-bit activations.

Although ACT has already demonstrated impressive compression capabilities, previous work on ACT is restricted to specific NN architectures. 
For example, ActNN~\cite{chen2021actnn} is a quantization framework for convolutional NNs only; Mesa~\cite{pan2021mesa} proposes a per head/layer quantization method for vision transformers; and AC-GC~\cite{evans2021ac} derives convergence error bound for different types of operators separately. 

Developing a generic ACT framework is challenging. Theoretically, convergence guarantees must be made without assumptions on the network architecture. Algorithmically, the framework should find effective compression strategies for all kinds of networks automatically. From the system perspective, the framework should support arbitrary NN operations, including user-defined ones. 

%

In this work, we propose \method, a general framework for ACT that is agnostic to the NN architecture. 
Neither specialized mathematical derivations nor customized implementation is needed to support different operators.
To enable this, we develop a general convergence theory by analyzing the stochastic gradient (SG) introduced by ACT. We show that the SG can be well approximated by a linearized version, which is unbiased to stochastic compressors. The variance of the linearized gradient has a particularly simple structure that allows a numerical algorithm to predict the variance given a compression strategy. Then, we generate the strategy by approximately solving an integer program.

We implement our method as a library based on PyTorch that can be quickly integrated into real-world machine learning systems. 
The library also provides several optimization levels to explore the trade-off between memory and speed. 
We demonstrate the flexibility and efficiency of \method on various tasks, including image classification, object detection, text, and graph node classification. Our evaluation shows that \method can reduce activation memory by up to 8.1$\times$, enabling training with a 24.7$\times$ larger batch size on the same GPU. 
In sum, our main contributions are as follows:
\begin{itemize}
    \itemsep0em
    \item We propose a general convergence theory for ACT.
    \item We develop an algorithm that automatically estimates the sensitivity of each compressed tensor and selects the optimal compression strategy. 
    \item We build efficient implementation of \method in PyTorch with an easy-to-use API that can also be combined with other memory-saving techniques seamlessly.
\end{itemize}


%% file: 2-related.tex
\section{Related Work}\label{sec:related}

\textbf{Activation Compressed Training.} 
ACT has been applied to convolutional NNs using different compressors, such as quantizers~\cite{chakrabarti2019backprop, fu2020don,chen2021actnn}, JPEG~\cite{evans2020jpeg}, or scientific data compression algorithms~\cite{jin2021novel,evans2021ac}. 
ACT is also applied to transformers~\cite{pan2021mesa} and graph NNs~\cite{anonymous2022exact}.

However, the existing theory for ACT~\cite{chakrabarti2019backprop,fu2020don,chen2021actnn,evans2021ac} relies on the case-by-case analysis of specific network operators, such as convolution, ReLU, and batch normalization.
It also requires dedicated implementations for each operator. 
On the contrary, \method focuses on the generality of activation compressed
training, not a specific quantizer design, which is the main
topic of previous work. Instead of assuming that the network is a stack
of layers, GACT formulates the problem as a computational graph of operators. 
This is general enough to cover transformers~\cite{vaswani2017attention}, graph NNs~\cite{kipf2016semi}, second-order
derivatives, and unknown future~architectures.





\noindent{\textbf{Reduced Precision Training.}}
Apart from ACT, reduced precision training~\cite{micikevicius2018mixed, wu2018training,wang2018training,banner2018scalable,chen2020statistical,sun2020ultra} performs calculations directly on low precision data, reducing the computation cost and memory footprint simultaneously. To achieve this, specialized kernels are used to calculate on low precision data. In contrast, ACT only considers storage, and it can thus use more flexible compression strategies and achieve a much better compression ratio with the same accuracy loss.

\noindent{\textbf{Memory-Efficient Training.}}
Gradient checkpointing~\cite{chen2016training, jain2019checkmate} trades computation for memory by dropping some of the activations in the forward pass from memory and recomputing them in the backward pass.
Swapping~\cite{kirisame2020dynamic, huang2020swapadvisor, wang2018superneurons, peng2020capuchin} offloads activation or model parameters to an external memory (e.g., CPU memory).
Recent work~\cite{beaumont2021efficient} explores the possibility of combining the gradient checkpointing and swapping.
All these methods save memory by storing fewer tensors on the GPU. In contrast,
\method compresses the saved tensors and is complementary to these approaches. Moreover, the generality of \method enables easy combination with these methods, which we explore in this paper.

%% file: 3-theory.tex

\section{Formulation}

We first present the mathematical formulation of our activation compressed training (ACT) framework. 
As we would like to develop a general ACT algorithm, applicable to a wide range of NN architectures, we make minimal assumptions on our formulation. 
Throughout the paper, we define the variance of a vector $x$ as $\Var{x}=\E{\norm{x}^2}-\norm{\E{x}}^2$.

\subsection{Activation Compressed Training}
In this work, we abstract the forward propagation as two functions $\ell(x; \theta)$ and $h(x; \theta)$.
Both take a datum $x$ and the model parameter $\theta$ as the input. The loss function $\ell(x; \theta)$ outputs the loss of the network $\theta$ on datum $x$. The context function $h(x; \theta)$ outputs tensors to be stored in the memory for computing the gradients, which are referred as the \emph{context}.   Assume that the context consists of $L$ tensors, where each tensor $h^{(l)}(x; \theta)$ is represented by a flattened $D_l$-dimensional vector. 
Denote $h(x; \theta)=(h^{(l)}(x; \theta))_{l=1}^L$. Our notations are somewhat unconventional in the sense that we do not explicitly define each layer's activation. 
We do not even assume that there is a NN. It could be any computational graph that saves context tensors.


Given a dataset $\Xc=\{x_n\}_{n=1}^N$, define the batch loss $\Lc(\theta):=\frac{1}{N}\sum_{n=1}^N \ell(x; \theta)$.
The dataset can be equivalently represented as an empirical data distribution $p_{\Xc}(x):=\frac{1}{N}\sum_{n=1}^N \delta(x-x_n)$, where $\delta$ is the Dirac delta function. The batch loss can be written as $\Lc(\theta)=\Eb_{\Xc}[\ell(x; \theta)]$, where $\Eb_{\Xc}$ denotes for taking expectation over $p_{\Xc}$.

The network is trained with stochastic gradient descent (SGD)~\cite{bottou2010large}. Starting from an initial model $\theta_0$, at the $t$-th iteration, SGD updates the model with:
\begin{align}
\theta_{t+1}\leftarrow \theta_t - \eta \nabla_{\theta} \ell(x; \theta_t),\label{eqn:sgd}
\end{align}
where $\eta$ is a learning rate, and the SG $\nabla_{\theta} \ell(x; \theta)$ is computed on a random datum $x\sim p_{\Xc}$. Notice that $\Es{\Xc}{\nabla_\theta \ell(x; \theta)}=\nabla_{\theta} \Lc(\theta)$, i.e., the SG is an unbiased estimator of the batch gradient $\nabla_{\theta} \Lc(\theta)$.

Crucially, the SG can be written in the form $\nabla_{\theta} \ell(x; \theta_t)=g(h(x; \theta_t); \theta_t)$. In other words, the back propagation only depends on the forward propagation through the context $h(x; \theta_t)$. 
The entire context must be kept in memory for computing the gradients. The context dominates the memory consumption in many applications. 

\begin{figure}[t]
	\centering
	\includegraphics[width=\mywidth]{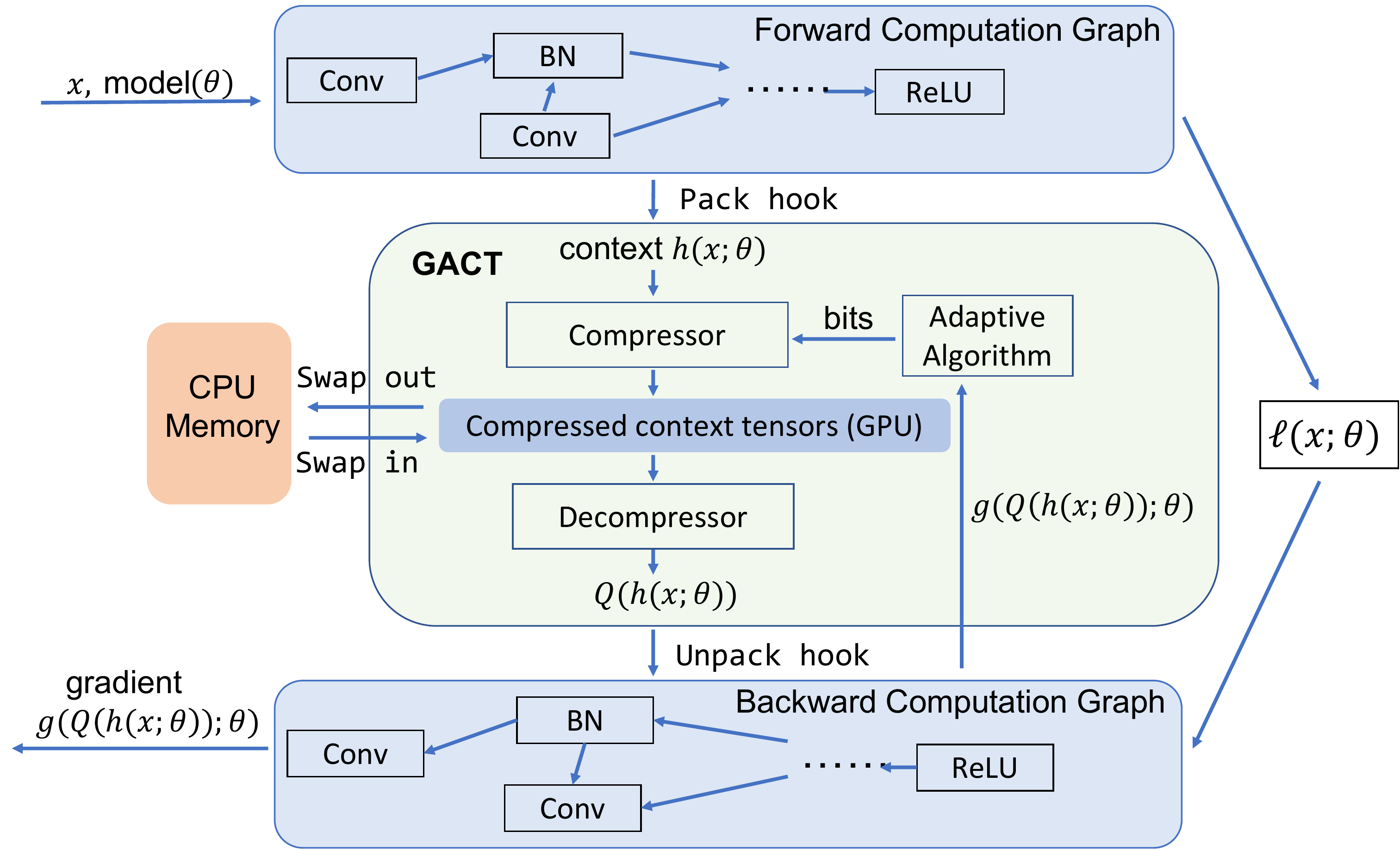}
	\ifisarxiv
	\else
	\vspace{-2em}
	\fi
	\caption{\small The architecture of \method.}
\label{fig:architecture}
\end{figure}

ACT reduces the training memory footprint by compressing the context. 
Let $Q(h)$ be a compressor, which converts $h$ to compact formats while keeping $Q(h)\approx h$.
Then, ACT computes the gradient with compressed context: \begin{align}\label{eqn:ac}
\theta_{t+1}\leftarrow \theta_t - \eta g(Q(h(x; \theta_t)); \theta_t).
\end{align}
We refer to $g(Q(h(x; \theta_t); \theta_t)$ as the activation compressed (AC) gradient. 
ACT is significantly more memory efficient then the plain SGD, Eq.~(\ref{eqn:sgd}), since it only needs to store a compressed version of the context. Suppose the original context $h(x; \theta_t)$ consists of 32-bit floating point tensors, and $Q(\cdot)$ is a compressor which quantizes tensors to 2-bit integers, ACT will reduce the context memory by $16\times$.
Fig.~\ref{fig:architecture} illustrates the computational graph of ACT with these notations. 
In the following presentation, we might denote $h(x, \theta)$ simply by $h$ when there is no confusion. 

\subsection{Convergence of ACT}\label{sec:convergence}
ACT is a lossy approximation of SGD, as it uses an approximate gradient $g(Q(h); \theta)$. Therefore, some kind of theoretical guarantee is required for ACT to be useful. 
Fortunately, analyzing ACT is made significantly simpler by introducing an \emph{unbiased stochastic} compressor $Q(\cdot)$, such that $\Es{Q}{Q(x)}=x$ for any $x$. $\Es{Q}{\cdot}$ means taking expectation over the compressor. 
In this way, $g(Q(h); \theta)$ can be viewed as a stochastic estimator of the batch gradient $\nabla \Lc(\theta)$, but the randomness comes not only from the datum $x$ but also the compressor $Q(\cdot)$. Therefore, ACT is still an SGD algorithm.
Standard analytical tools for SGD~\cite{bottou2018optimization} are applicable for studying ACT. 

SGD algorithms have particular good properties when the SG is unbiased. In our case, this means $\Es{Q}{g(Q(h);\theta)}=g(h;\theta)$. However, the SG is biased general, even when the stochastic compressor itself is unbiased.\footnote{Consider the example $g(h)=\Ib(h\ge 0.5)$, where $h\in [0, 1]$ and its AC gradient $g(Q(h))=\Ib(Q(h)\ge 0.5)$  with the compressor $Q(h)\sim \mathrm{Bernoulli}(h)$. Then, $\E{g(Q(h))}=P(Q(h)=1)=h\ne g(h).$
} 

The key technique in this work is to construct an unbiased approximation of the AC gradient by linearizing the gradient function $g(\cdot; \theta)$. Consider the first-order Taylor expansion of $g(\cdot; \theta)$ at $h$:
\begin{align}\label{eqn:first-order}
\hat g(Q(h); h, \theta):=
g(h; \theta) + 
J(h, \theta)
\Delta h,
\end{align}
where $J(h, \theta):=\frac{\partial g(h;\theta)}{\partial h}$ is a Jacobian matrix, $\Delta h:=Q(h)-h$ is the compression error. 
We further denote $\hat g_{x\theta}(Q(h); h):=\hat g(Q(h); h, \theta)\vert_{h=h(x; \theta)}$ and $J_{x\theta}(h):=J(h, \theta)\vert_{h=h(x; \theta)}$ for short. 
Since $\E{\Delta h(x; \theta)}=0$, 
$\hat g_{x\theta}(Q(h); h)$ is an unbiased SG,  Furthermore, the approximation error is small:
\begin{proposition}\label{prop:bias-order}
Assuming that $g(h; \theta)$ is twice differentiable w.r.t. $h$, and the second order derivative is bounded, then 
\begin{align*}&\E{\norm{g(Q(h); \theta) - \hat g_{x\theta}(Q(h); h)}_2}
= O(\Vars{Q}{\Delta h}).\end{align*}
\end{proposition}

Since $\Delta h$ itself is unbiased,  $\Vars{Q}{\Delta h}=\Es{Q}{\norm{\Delta h}^2}$ is simply the expected compression error. Prop.~\ref{prop:bias-order} implies that the linearization error is bounded by the compression error. The linearized gradient $\hat g$ is accurate if the compression is accurate.  Using $\hat g$ as a bridge, we arrive in the following convergence theorem:
\begin{theorem}\label{thm:convergence}
	Assume that:\\
\noindent\textbf{A1.} $\Lc(\theta)$ is a continuous differentiable, $\nabla\Lc(\theta)$ is $\beta$-Lipschitz continuous.\\
\noindent\textbf{A2.} $\Lc(\theta)$ is bounded below by $\Lc_*$.\\
\noindent\textbf{A3.} $g(h; \theta)$ is differentiable w.r.t. $h$ and $\exists b>0$, s.t. $\forall \theta, \Eb\norm{g(Q(h(x; \theta)); \theta)-\hat g_{x\theta}(Q(h); h)}\le b$.\\
\noindent\textbf{A4.} $\exists \sigma^2>0$, s.t.,  $\forall\theta$, $\Var{\hat g_{x\theta}(Q(h); h)}\le \sigma^2$. \\
Then, for all $\eta < \frac{1}{2\beta}$, if we run ACT defined as Eq.~(\ref{eqn:ac}) for $T$ iterations, then we have

\begin{align*}
\min_{t=0, \dots, T-1}\E{\norm{\nabla \Lc(\theta_t)}^2}
\le \frac{4(\Lc(\theta_0)-\Lc_*)}{\eta T}+3b^2+ \eta\beta\sigma^2
\end{align*}

\end{theorem}

\paragraph{Remark: } The analytical technique used in Thm.~1 is rather standard, see Thm.~4.8 in~\citet{bottou2018optimization}. However, we consider the variance term $\sigma^2$ of the \emph{linearized gradient}, rather than the SG itself. This formulation brings better analytical properties and an adaptive algorithm for determining the compression scheme, as we shall see soon in Sec.~4.

The convergence of ACT is affected by both the linearization error (A3) and the variance of the unbiased gradient $\hat g(\cdot; \theta)$ (A4). The latter is characterized as:

\begin{proposition}\label{prop:var-order}
$
\Var{\hat g_{x\theta}(Q(h); h)}
=\Vars{\Xc}{g(h; \theta)} + 
 \Eb_{\Xc}\left[\Vars{Q}
{
\hat g_{x\theta}(Q(h); h)
}\right],
$
where the second term on the RHS equals to 
$
 \Eb_{\Xc}\left[\Vars{Q}
{J_{x\theta}(h)\Delta h}\right]=O\left(\Vars{Q}{\Delta h}\right).$
\end{proposition}

Prop.~\ref{prop:var-order} separates the variance from different noise sources. 
$\Vars{\Xc}{g(h(x, \theta); \theta)}$ is the variance raised by random sampling of data (``sampling variance''). 
$\Eb_{\Xc}\left[\Vars{Q}
{J_{x\theta}(h)\Delta h(x, \theta)}\right]$ is the variance raised by compression. Now, the convergence in Thm.~\ref{thm:convergence} is depicted by $3b^2+\eta\beta\sigma^2$. By Prop.~\ref{prop:bias-order}, $b^2=O(\Vars{Q}{\Delta h}^2)$. By Prop.~\ref{prop:var-order}, $\sigma^2 =O(1)+O(\Vars{Q}{\Delta h})$, since the sampling variance is not affected by compression. Therefore, when the compression is accurate ($\Delta h\rightarrow 0$), the impact of the linearization error is negligible, and the variance of the unbiased gradient dominates. ACT behaves as if the AC gradient is unbiased.

%% file: 4-adaptive.tex
\section{Adapting the Compression Rate}

In a network, some context tensors (such as those stored for computing the cross entropy loss) are extremely sensitive, a small amount of compression would result in diverged training, while other tensors are quite robust to compression. 
Therefore, we must apply different amounts of compression for each context tensor. 
As a general framework, we have no prior knowledge of the users' model architecture, so we designed an algorithm to infer the sensitivity for each context tensor and determine their compression rate automatically. 

There is a tradeoff between the compression error and the storage requirement.
We represent the storage requirement of the compressed context in \emph{bits per dimension}. We assume that $b_l$ bits/dim. are used for compression $h^{(l)}$, and $Q_{b_l}(h^{(l)})$ be the compression result. Let $b=(b_l)_{l=1}^L$ be a \emph{compression scheme}, $Q_b(h):=\{Q_{b_l}(h^{(l)})\}_{l=1}^L$, and $\Delta_b h = Q_b(h) - h$.

\subsection{Structure of Variance}

As discussed in Sec.~\ref{sec:convergence}, when the compression is relatively accurate, the variance plays the main role in determining the convergence. Therefore, we would like to investigate how the compression scheme would impact the variance. Formally, we are interested in: 
\begin{align*}
V(b; h, \theta) := \Vars{Q}
{
\hat g(Q_b(h); h, \theta)
}.
\end{align*}
Once $V(b, h; \theta)$ is known, we can find the minimum variance compression scheme under a given total bits budget $B$, by solving the integer programming problem:
\begin{align}
\min_b V(b; h(x; \theta), \theta),~~~\mbox{s.t. }\sum_{i=1}^L b_l D_l \le B,\label{eqn:ilp}
\end{align}
where $D_l$ is the dimensionality of $h^{(l)}$. 
To proceed, we need the following assumptions on the compressor $Q_b(\cdot)$:\\
\noindent\textbf{Assumption B1: } The compressed result is element-wise uncorrelated. That is, for any $i\ne j$, $\Cov{Q_b(h)_i}{Q_b(h)_j}=0$.\\
\noindent\textbf{Assumption B2: } For compressing $h^{(l)}(x; \theta)$ to $b_l$ bits/dim., the compression error can be written in the form $\Var{\Delta_{b_l} h^{(l)}(x; \theta)_j}\le  R_{lj}(x; \theta)S(b_l)$, where $S(b_l)$ is a known function. This isolates the effect of $b_l$ through the unary factor $S(b_l)$.\\
Both assumptions can be achieved by a stochastic rounding quantizer~\cite{courbariaux2015binaryconnect}, where $R_{lj}(x; \theta)=\frac{1}{4}\left(\max_k h_k^{(l)}-\min_k h_k^{(l)}\right)^2$ and 
$S(b) = (2^{b_l}-1)^{-2}$. See Appendix~\ref{sec:appendix-var-structure} for the derivations. 

The following theorem reveals the structure of the variance:
\begin{theorem}\label{thm:variance-structure}
Under assumptions B1, B2, there exists a family of functions $\{c_l(h, \theta)\}_{l=1}^L$, such that the compression variance can be written in the form
\begin{align}\label{eqn:var-decomposition}
V(b; h, \theta)\le \sum_{l=1}^L c_l(h, \theta) S(b_l).
\end{align}
\end{theorem}

\subsection{Computing Sensitivity}
\label{sec:compute_sensitivity}
Thm.~\ref{thm:variance-structure} reveals two good properties of the variance: (1) the impact of compressing different context tensors simply sums up, without affecting each other; and (2) the compression scheme only impacts the variance through $S(b_l)$. Both properties are brought about by linearization. Since $S(\cdot)$ is a known function, we only need to know $c_l(h, \theta)$ to solve  problem Eq.~(\ref{eqn:ilp}). 
$c_l(h, \theta)$ can be understood as the sensitivity of the AC gradient to the compression of the $l$-th tensor.
We can compute $c_l(h, \theta)$ numerically
by leveraging the idempotence of compressing a tensor:
\\
\noindent\textbf{Assumption B3: } If $h=Q(h^\prime)$ for some $h^\prime$ with non-zero probability, then $Q(h)=h$ and $\Vars{Q}{Q(h)}=0$.

Let $Q^{\neg (l)}_b(h)=\{Q_{b_1}(h^{(1)}), \dots, h^{(l)}, \dots, Q_{b_L}(h^{(L)})\}$ be some tensors, where every tensor except $h^{(l)}$ is compressed. Plug $h=Q^{\neg (l)}_b(h)$ into Eq.~(\ref{eqn:var-decomposition}), and use B3, we have
\begin{align*}
V(b; Q^{\neg (l)}_b(h), \theta) \le c_l(Q^{\neg (l)}_b(h), \theta) S(b_l).
\end{align*}
The left hand side can be approximated by taking $\hat g(Q_b(h); h, \theta)\approx g(Q_b(h); \theta)$. Assume that $c_l(\cdot, \theta)$ is reasonably continuous, we have
\begin{align*}
c_l(h, \theta) \approx \Vars{Q}
{
g(Q_b(h); \theta)
}\vert_{h=Q_b^{\neg (l)}(h)} / S(b_l).
\end{align*}
The variance can be replaced by empirical variance.

\begin{algorithm}[t]
\caption{Numerical algorithm for computing $c_l(h, \theta)$.}\label{alg:autoprec}
\begin{algorithmic}
\REQUIRE A gradient evaluation function $g(\cdot; \theta)$
\REQUIRE A series of $L+1$ random seeds $(r_l)_{l=1}^{L+1}$.
\REQUIRE Any compression scheme $b=(b_l)_{l=1}^L$
\STATE $\forall l$, seed $Q^{(l)}$ with $r_l$
\STATE $g_0\leftarrow g(Q_b(h); \theta)$ \COMMENT{First iteration}
\STATE $\forall l$, seed $Q^{(l)}$ with $r_l$
\STATE seed $Q^{(l)}$ with $r_{L+1}$
\STATE $g_1\leftarrow g(Q_b(h); \theta)$ \COMMENT{Second iteration, with another seed}

\textbf{Return} $\frac{1}{2}\norm{g_0 - g_1}^2 / S(b_l)$
\end{algorithmic}
\end{algorithm}

Alg.~\ref{alg:autoprec} illustrates this idea. To compute $\Vars{Q}
{
g(Q_b(h); \theta)
}$ at $h=Q_b^{\neg (l)}(h)$, we keep the random seeds fixed for all the compressors except the $l$-th one. We compute the empirical variance by two evaluations of $g(Q_b(h); \theta)$, which are two NN iterations (forward + backward propagation). 

Finally, we assume that $c(h, \theta)$ remains stable for different mini-batches $h$, and along the training trajectory $(\theta_t)$. Therefore, we maintain a $c_l$ for each tensor $l$, which is updated by periodically running Alg.~\ref{alg:autoprec}. 
Eq.~(\ref{eqn:ilp}) is approximately solved by the $O(L\log_2 L)$ greedy algorithm~\cite{chen2021actnn}.

Another useful feature of this approach is predicting failure (in an \emph{a posteriori} manner). If the compression variance $V(b; h, \theta)$ is dominating the overall gradient variance $\Var{g(Q(h); \theta_t)}
$, compression is adding too much noise to the gradient, and the convergence might be affected. The overall gradient variance can be computed by maintaining a running mean of the gradient. If $V(b; \theta)/\Var{\hat g(Q(h); \theta_t)}$ is too large, we can raise an alert to the user to increase the storage budget. 

%% file: 5-system.tex
\ifisarxiv
\else
\vspace{-1em}
\fi
\section{System Implementation}

We implemented \method as a lightweight library in PyTorch. Users can use \method for any NN architecture with several lines of code change.
\method uses low-level PyTorch hooks to capture context tensors, so it supports arbitrary operators, including custom operators defined by users.
We implemented efficient CUDA kernels to infer tensor sensitivity and to perform compression during run time.
\method uses the same per-group quantizer in ActNN~\cite{chen2021actnn} as the compressor. However, \method differs from ActNN in several aspects.
ActNN relies on manual analytical deduction to compute the sensitivity for different operators, while \method infers tensor sensitivity automatically, as described in Sec.~\ref{sec:compute_sensitivity}.
Moreover, ActNN performs layer-level quantization.
It has to implement an activation compressed version for each operator and substitute operators during the training (e.g., replace {\code torch.nn.Conv2d} with {\code actnn.Conv2d}).
In contrast, \method runs at tensor level and uses a single hook interface to compress saved tensors for all operators.

\subsection{General API}
\ifisarxiv
\begin{figure}[t]
	\centering
	\includegraphics[width=0.5\linewidth]{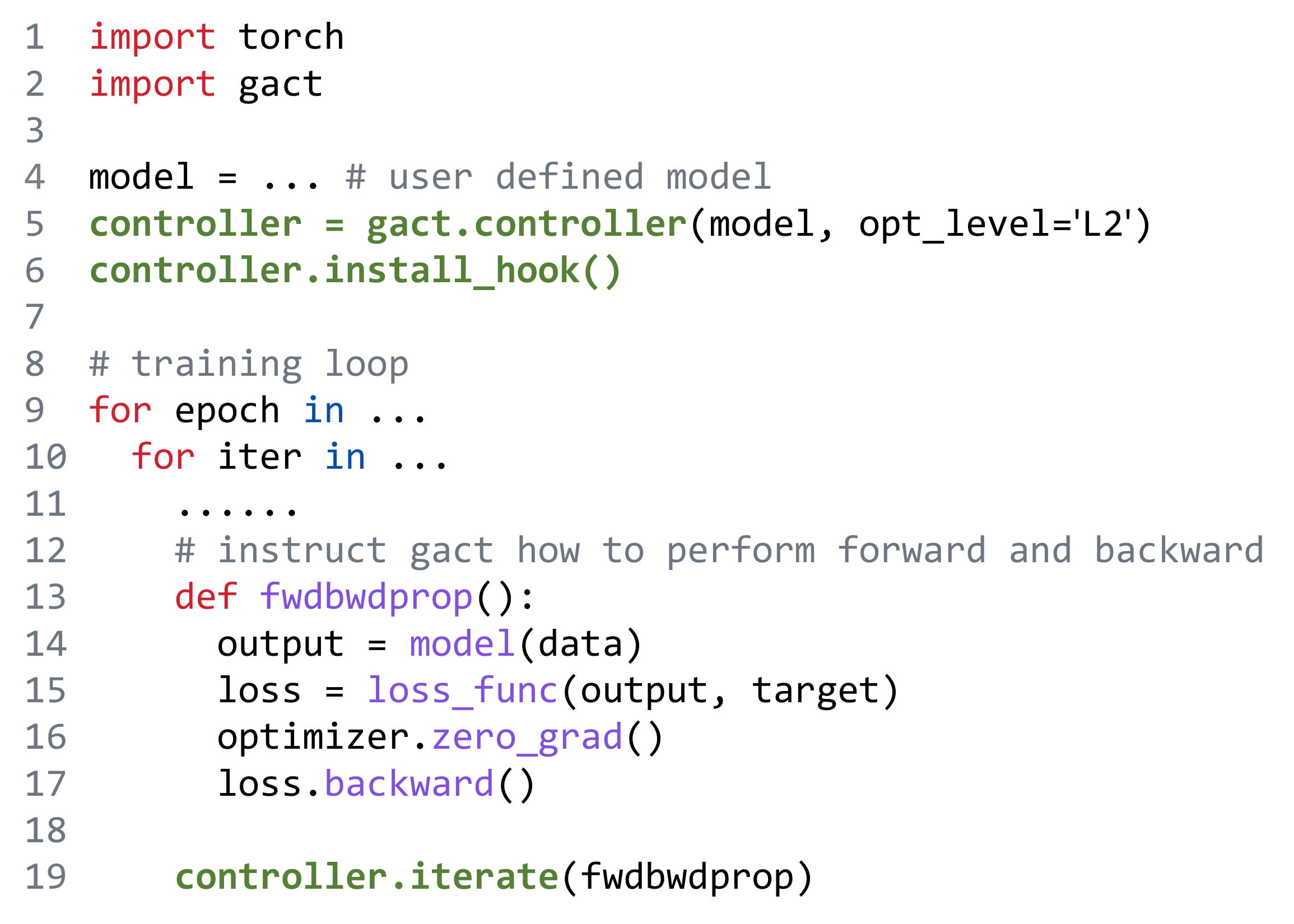}
	\caption{Usage example of \method} 
\label{fig:act_layers}
\end{figure}
\else
\begin{figure}[t]
	\centering
	\includegraphics[width=0.92\linewidth]{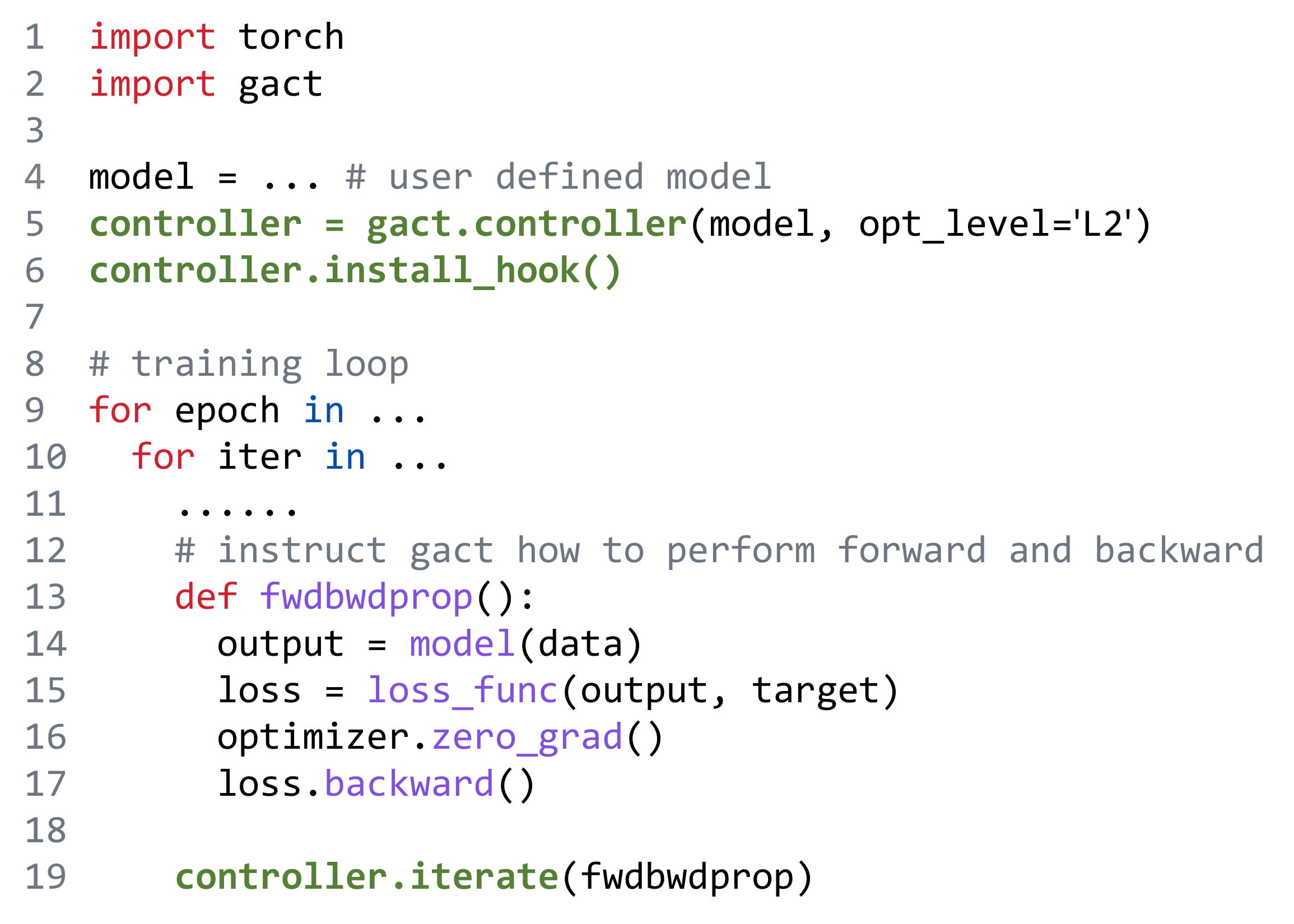}
	\vspace{-1.4em}
	\caption{\small Usage example of \method} 
\label{fig:act_layers}
\end{figure}
\fi

As shown in Fig.~\ref{fig:act_layers}, the interface of \method is straightforward and intuitive, requiring the user to (i) initialize the \method controller and specify an optimization level (Line 5); (ii) install hooks (Line 6); and (iii) instruct \method
how to perform forward and backward propagation (Lines 13-17) and pass it as a function ({\code fwdbwdprop}) to the controller (Line 19). We require users to specify (iii) because \method needs to numerically run the forward and backward pass to infer tensor sensitivity. Although {\code fwdbwdprop}
is passed to the controller every iteration, it is only called internally every {\code adapt\_interval} iterations when tensor sensitivity changes. As shown in Sec.~\ref{sec:cpr_strategy}, tensor sensitivity stabilizes quickly after the first several epochs, {\code adapt\_interval} can thus be set to a large number, introducing negligible impact on training speed.


\subsection{System Architecture}
Fig.~\ref{fig:architecture} shows an overview of \method. The \method controller has three modules: Adaptivate Algorithm; Compressor; and Decompressor. In the forward pass, the controller uses PyTorch {\code pack\_hook} to capture all context tensors. Then Adaptive Algorithm infers tensor sensitivity based on gradients and assigns higher bits to more sensitive tensors, as described in Sec.~\ref{sec:compute_sensitivity}. The bits information is used to instruct Compressor to perform quantization. In the backward pass, Decompressor dequantizes context tensors and uses {\code unpack\_hook} to send the dequantized results back to the PyTorch's auto differentiation engine. The controller is also responsible for swapping quantized tensors to the CPU and prefetching them back during the backward propagation if swapping is enabled.

\subsection{Identifying Tensors to Quantize} The {\code pack\_hook} and {\code unpack\_hook} process all types of context tensors, including activation, parameters trained by the optimizer, and training states such as running mean/variance used by batch normalization. To guarantee that only the activations are quantized, we filter out saved parameters by recording the data pointers of all the model parameters before training, and we skip quantization if the input tensor pointer exists in the parameter pointer set. Similarly, \method does not quantize training states by checking if the input tensor requires gradients. 

However, using hooks blindly disables some memory-saving optimization. For example, 
in a transformer's self-attention layer,
the keys, query, value tensors are all calculated from the same input tensor. The saved objects of the three operations thus all refer to the same tensor.
In this case, PyTorch triggers the {\code pack\_hook} three times. If we perform quantization blindly, we waste computation resources and introduce extra memory consumption because the same underlying tensor is quantized and saved more than once. \method avoids duplication by generating footprints for each input context tensor. We use the CUDA data pointer, sampled data points, and the tensor statistics (e.g., sum) as the footprint. \method manages all quantized context tensors and uses the footprint to differentiate them. If a tensor is already quantized, \method will skip quantization and return previous results directly.



\subsection{Parallel Swap and Prefetch}
To further reduce activation memory, we combine \method with swapping. All compressed tensors are offloaded to the CPU during the forward pass and swapped back in the backward pass. Here, we replace the original tensor with quantized activation, as data movement is more expensive than computation. Swapping the original tensor saves the quantization overhead but adds more data movement cost between CPU and GPU. As shown in Sec.~\ref{sec:swap}, quantization overhead is much smaller than copying full-precision data to CPU in modern GPU architecture.

Furthermore, we create two new streams (swap in/out) to parallelize the computation and swapping operation to reduce the swap overhead. The forward computation and swap-out process happen in parallel during the forward pass. During the backward pass, in each layer the swap-in stream is responsible for prefetching the compressed activation of the previous layer to avoid synchronization overhead. 
We leverage the CUDA event to ensure tasks in different streams are executed in the correct order.

\subsection{Other Memory Optimizations}
\textbf{Gradient checkpointing.} 
Gradient checkpointing~\cite{chen2016training} works by dividing the NN into segments.
The algorithm only stores the inputs of each segment and recomputes the dropped activations segment by segment during backpropagation.
The memory consumption is thus the cost of storing the inputs of all segments plus the maximum memory cost to backpropagate each segment.
When combined with gradient checkpointing, \method can reduce the memory consumption of both parts.
\method reduces the memory consumption of the first part by quantizing the segment inputs. Moreover, the activations saved during the recompute phase are also quantized, reducing the memory cost of the second part.
Combining \method with gradient checkpointing might introduce more training noise because the recompute starts from quantized segment inputs, making the forward pass of recompute phase not exact. However, in Sec.~\ref{sec:swap}, we show the noise introduced by forwarding from the quantized tensors is negligible.

\ifisarxiv
\noindent{\textbf{Memory efficient self-attention.}}
\else
\textbf{Memory efficient self-attention.}
\fi
When the batch size is very large, the single layer after dequantization occupies a large amount of memory and prevents the batch size from increasing further. We observe this problem in transformer-based models where self-attention has quadratic space complexity in terms of sequence length.
To reduce the memory footprint of the self-attention layer, we implement the algorithm introduced in~\cite{rabe2021self} that achieves linear space complexity, and combines it with \method. 

\subsection{Optimization level}
\begin{table}[t]
	\centering
	\vspace{-1em}
	\caption{\small Optimization levels for \method. \label{tab:opt_level}}
	\resizebox{0.9\mywidth}{!}{
		\begin{tabular}{ccc}
			\toprule
			Level &  Compression Strategy & Bits\\
			\midrule
			L0   & Do not compress & 32\\
			L1   & per-group quantization with auto-precision & 4\\
			L2   & L1 + swapping/prefetching & 4 \\
			CB1  & L1 + gradient checkpointing & 4\\
			CB2  & CB1 + efficient self-attention & 4\\
			\bottomrule
	\end{tabular}}
\end{table}



To exploit the trade-off between memory saving and training speed, \method provides several optimization levels.
Higher levels can save more memory but with more overhead.
Tab.~\ref{tab:opt_level} lists these optimization levels. 
L1 uses per-group quantization with the adaptive algorithm. L2 combines per-group quantization with swapping and prefetching.
For transformer-based models, CB1 combines \method with gradient checkpointing. CB2 further reduces the peak memory by adding efficient self-attention to CB1.

%% file: 6-experiments.tex
\section{Experiments}
We first demonstrate the effectiveness of the \method adaptive algorithm. We further apply \method to a wide range of machine learning tasks, including image classification, object detection, text, and graph node classification. We compare the training accuracy and activation compression rate for full precision, adaptive 4/3/2 (using \method to adaptively decide quantization bits with an average of 4/3/2 bit) and fix-4 bit (quantizating all tensors uniformly with 4 bits).
Next, we study the trade-off between compression rate and training throughput and compare \method with other state-of-the-art memory-saving methods. Lastly, we demonstrate the flexibility of \method by exploring the possibility of combining it with other memory optimization methods (CB1, CB2 as listed in Table~\ref{tab:opt_level}). We use open-source model implementations for all tasks.

\subsection{Compression Strategy}
\label{sec:cpr_strategy}

\ifisarxiv
    \begin{figure}[t]
    \centering 
    \subfigure[Inferred per-tensor $c_l$ (line) and bits/dim. (bar) for VGG-11. Layers with * have a preceding ReLU layer with shared context. drop=dropout, loss=cross entropy~loss. ]{
    \includegraphics[width=0.36\linewidth]{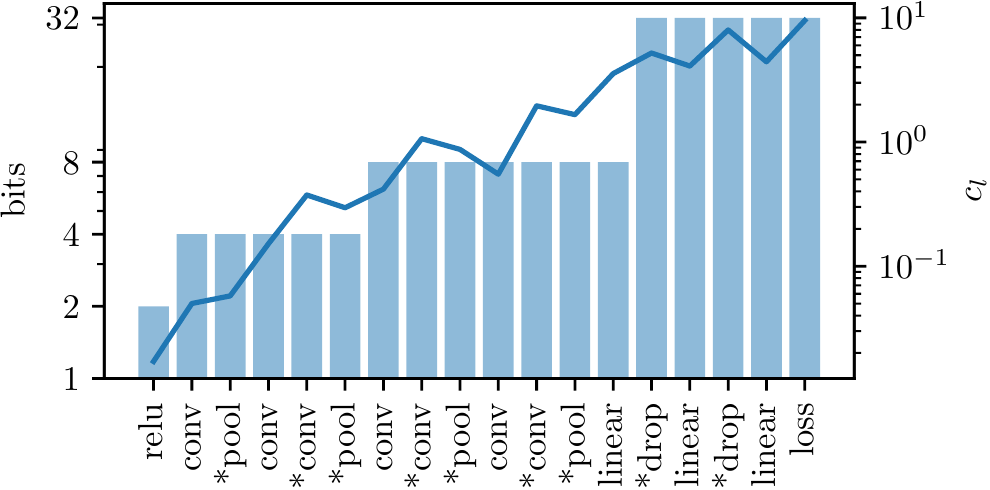}
    }
    \subfigure[Gradient variance.]{
    \includegraphics[width=0.18\linewidth]{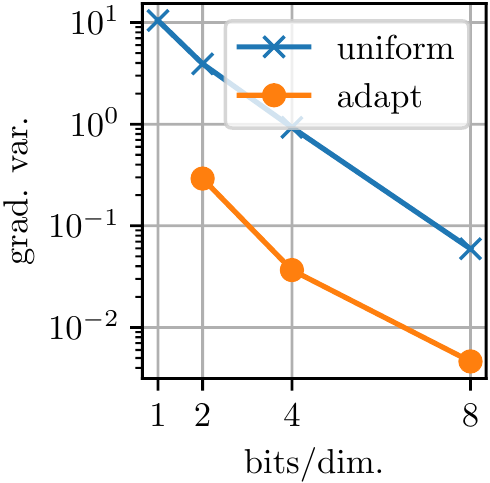}
    }
    \subfigure[Evolution of the per-tensor sensitivity. Each line is $c_l$ for a tensor.]{
    \includegraphics[width=0.18\linewidth]{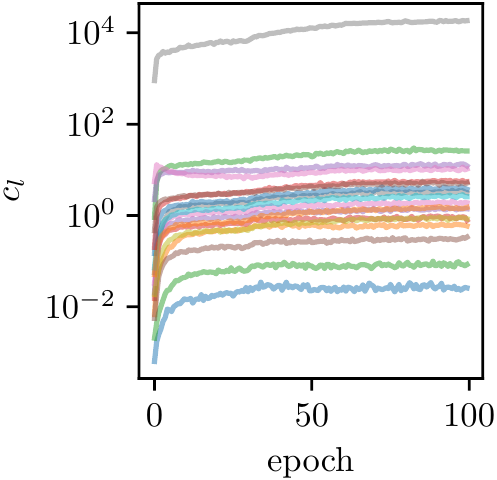}
    }
    \caption{Effectiveness of the adaptive algorithm.}
    \label{fig:sensitivity}
    \end{figure}
\else
    \begin{figure}[t]
    \centering 
    \subfigure[Inferred per-tensor $c_l$ (line) and bits/dim. (bar) for VGG-11. Layers with * have a preceding ReLU layer with shared context. drop=dropout, loss=cross entropy~loss. ]{
    \includegraphics[width=0.82\mywidth]{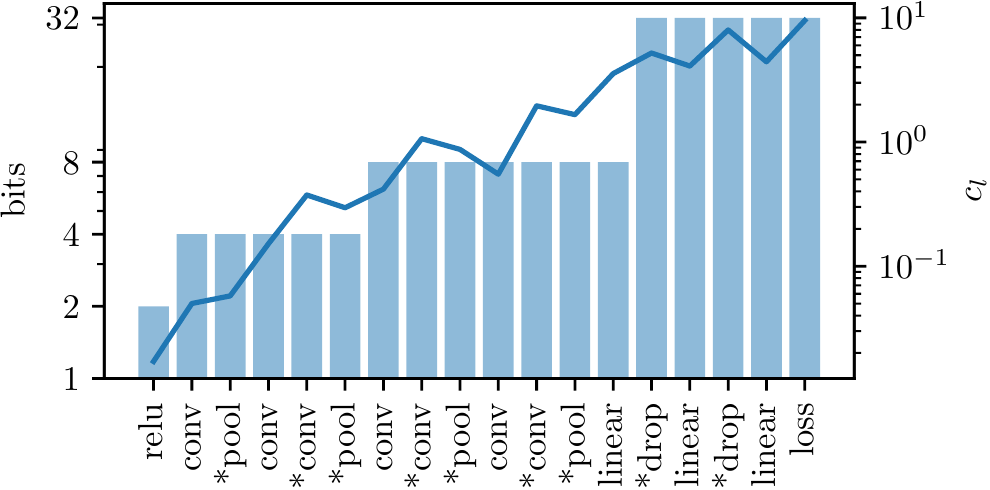}
    }
    \subfigure[Gradient variance.]{
    \includegraphics[width=0.4\mywidth]{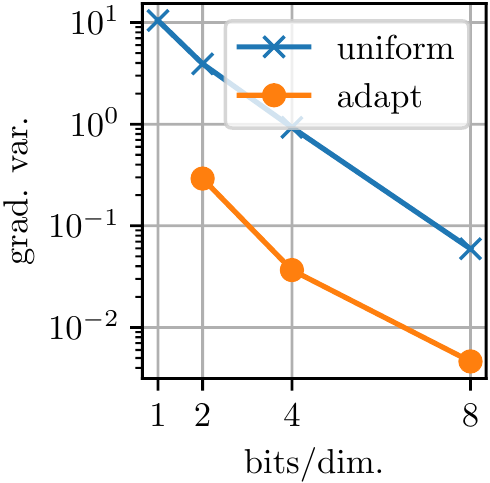}
    }
    \subfigure[Evolution of the per-tensor sensitivity. Each line is $c_l$ for a tensor.]{
    \includegraphics[width=0.4\mywidth]{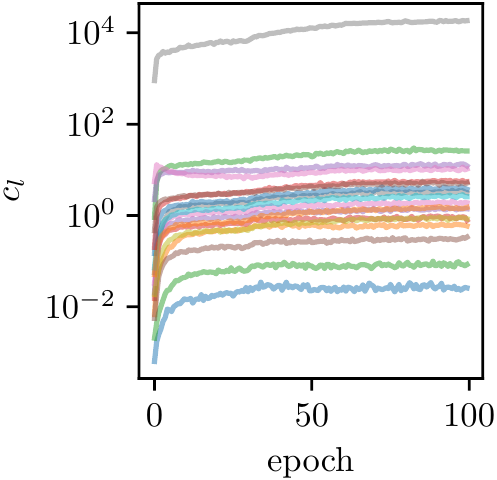}
    }
    \vspace{-1em}
    \caption{Effectiveness of the adaptive algorithm.}
    \label{fig:sensitivity}
    \end{figure}
\fi

We first test the effectiveness of our adaptive compression rate algorithm for training VGG-11~\cite{simonyan2014very} on ImageNet. Fig.~\ref{fig:sensitivity}(a) plots the inferred per-tensor sensitivity $c_l$ and the corresponding optimal bits/dim. \method assigns more bits to more sensitive layers. The context tensor saved by the cross-entropy loss operator is most sensitive. A small amount of compression leads to a huge gradient variance. This makes sense since the loss is the first operator to back-propagate through, where the error accumulates. Therefore, \method assigns 32 bits/dim. for the tensors in the classification head. With the adaptive algorithm, \method with an average of 4 bits/dim. achieves smaller gradient variance than uniformly assigning 8 bits/dim. for all the tensors, as shown in Fig.~\ref{fig:sensitivity}(b). Finally, Fig.~\ref{fig:sensitivity}(c) shows that the  sensitivity $c_l(h; \theta_t)$ remains stable during training.Therefore, periodically updating $c_l$ at a large interval is reasonable, and this introduces negligible impact on training speed.

\subsection{Optimization level}

\begin{table}[t]
    \centering
    \caption{For classification, we train VGG11~\cite{simonyan2014very}, ResNet-50~\cite{he2016deep}, and Swin-Tiny~\cite{liu2021Swin} on ImageNet~\cite{imagenet_cvpr09}. 
    For object detection, we train RetinaNet~\cite{lin2017focal}, Faster R-CNN~\cite{ren2015faster} on Coco~\cite{lin2014microsoft}. We report accuracy on validation sets (Div. indicates diverge) and the compression rate of context tensors (numbers in brackets) for both tasks. }
    \resizebox{\mywidth}{!}{
    \begin{tabular}{c|c|c|c|c}
    \toprule
    Task  &  Model &  FP32 &  \shortstack{\method \\Adapt 4bit (L1)} & \shortstack{\method \\Adapt 2bit} \\
    \midrule
    \multirow{2}{3em}{Cls.} 
    & VGG11     &   68.75  & 68.77 (2.84$\times$) & 68.49  (3.34$\times$) \\
    & ResNet-50 &   77.29  & 76.96 (6.69$\times$) & 76.13 (11.39$\times$)\\
    & Swin-tiny &   81.18  & 80.92 (7.44$\times$) & 77.91 (13.73$\times$)\\
    \hline
    \multirow{2}{3em}{Det.}
    & Faster RCNN & 37.4   & 37.0 (4.86$\times$) & 36.1 (6.81 $\times$)\\
    & RetinaNet   & 36.5   & 36.3 (3.11$\times$) & Div. \\
    \bottomrule
    \end{tabular}}
    \label{tab:vision_acc}
\end{table}

We apply \method on various computer vision tasks, including image classification and object detection, as shown in Fig.~\ref{tab:vision_acc}. We also vary the average bits used by the adaptive algorithm to explore the memory accuracy trade-off. On both tasks, \method L1 achieves comparable ($<0.5\%$ accuracy drop) or even better results than the full precision training, while reducing activation memory by up to 7.44$\times$. Here, we list the accuracy of FP32 as the strongest accuracy baseline. For other lossy methods we consider in Sec.~\ref{sec:mem-speed}, the accuracy is no better than FP32, and we list their training accuracy in Appendix~\ref{sec:baseline-acc}.
Notice that here \method Adapt 2bit diverges on the detection task. This is because, as 
shown in Sec.\ref{sec:convergence}, although ACT has unbiased gradients, the compression
error and learning rate affect the convergence. When using
2 bit, the compression error is large and the learning rate has to
be reduced accordingly to guarantee convergence. However,
we do not want to slow training by decreasing the learning
rate. All experiments are run with the same learning rate as
the full precision. Therefore when compression error is large,
the training diverges. 
Furthermore, we observe that the memory reduction varies among networks because \method does not quantize intermediate states, and the size of intermediate states differs between networks. For example, in VGG11, when the batch size is 128, \method reduces the saved tensor size from 5889MB to 2080MB, among which 78\% (1494MB) is used to store the intermediate index for the max-pooling layer that is not quantized by \method.

Next, we demonstrate the flexibility of \method by applying it to a wider variety of natural language processing (NLP) and graph machine learning (Graph) tasks. We run multiple seeds for each task, and we report the mean$\pm$std of accuracy/F1 across runs as shown in Tab.~\ref{tab:nlp-graph-acc-cpr}. We include the detailed experimental setup in Appendix \ref{sec:setup}. For both NLP and Graph tasks, \method L1 achieves comparable training results with FP32, introducing less than 0.3\% accuracy/F1-score drop, while reducing activation memory by 4.18$\times$ to 7.93$\times$. Moreover, the results are stable across runs, introducing similar accuracy variance as FP32.
We also show the training results of fix-4bit quantization, where all tensors are uniformly quantized with 4 bits. As shown in Tab.~\ref{tab:nlp-graph-acc-cpr}, fix-4 bit quantization causes significant accuracy/F1-score loss on various graph models. For Bert-large, fixed-4 bit quantization works fine because all the context tensors have similar sensitivity. 
On the other hand, \method L1, using a similar amount of memory as always quantizing each layer to 4 bits, still performs on par with full precision training on all the models. 
This shows the necessity of using adaptive algorithms to assign bits based on tensor sensitivity for stabilized training.
Moreover, for Bert-large and three graph models (GCN/GAT/GCNII), \method converges and gives lossless results with 3 bits. Remarkably, across all the graph models, training with 2-bit \method causes little accuracy loss ($<1\%$).
This shows the robustness of our adaptive algorithm.

\begin{table*}[t]
\ifisarxiv
\else
\vspace{-1em}
\fi
\caption{Accuracy and activation compression rate for NLP and Graph tasks. Accuracy that drops $>$ 1\% is in italic font.}
\label{tab:nlp-graph-acc-cpr}
\centering
\resizebox{0.96\linewidth}{!}{
\begin{tabular}{ c|c|c|c|c|c|c } 
\toprule
Model & Dataset & FP32 & Fix 4bit & \method Adapt 4bit (L1) &  \method Adapt 3bit & \method Adapt 2bit \\
\midrule
\multirow{4}{3em}{GCN} 
& Flickr & 51.17 $\pm$ 0.19    & 50.93 $\pm$ 0.16 (7.56$\times$)           & 51.08 $\pm$ 0.18 (7.93$\times$) & 51.14 $\pm$ 0.18 (11.34$\times$) & 51.20 $\pm$ 0.18 (17.56$\times$)\\
& Reddit & 95.33 $\pm$ 0.07    & 94.42 $\pm$ 0.11 (7.55$\times$)           & 95.32 $\pm$ 0.07 (7.90$\times$) & 95.31 $\pm$ 0.07 (9.70$\times$)  & 95.34 $\pm$ 0.06 (13.68$\times$)\\
& Yelp   & 39.86 $\pm$ 0.94    & 39.85 $\pm$ 1.22 (5.94$\times$)           & 40.06 $\pm$ 0.74 (6.42$\times$) & 40.21 $\pm$ 0.82 (7.46$\times$)  & 39.89 $\pm$ 1.45 (9.00$\times$)\\
& ogbn-arxiv & 71.51 $\pm$ 0.65 & \textit{68.61 $\pm$ 0.77 (7.54$\times$)} & 71.35 $\pm$ 0.36 (8.09$\times$) & 70.82 $\pm$ 0.95 (10.45$\times$) & 70.87 $\pm$ 0.66 (13.75$\times$)\\
\multirow{4}{3em}{GAT} 
& Flickr & 52.40 $\pm$ 0.28     & \textit{35.24 $\pm$ 11.90 (4.23$\times$)} & 52.26 $\pm$ 0.31 (4.34$\times$) & 51.68 $\pm$ 1.13 (5.04$\times$) & 51.62 $\pm$ 1.19 (5.46$\times$)\\
& Reddit & 95.95 $\pm$ 0.06     & \textit{59.37 $\pm$ 11.48 (4.12$\times$)} & 96.02 $\pm$ 0.09 (4.29$\times$) & 95.96 $\pm$ 0.06 (4.64$\times$) & 95.82 $\pm$ 0.06 (5.24$\times$)\\
& Yelp   & 52.41 $\pm$ 0.69     & \textit{36.09 $\pm$ 13.70 (4.04$\times$)} & 52.18 $\pm$ 0.38 (4.18$\times$) & 51.63 $\pm$ 0.83 (4.53$\times$) & 51.15 $\pm$ 0.53 (5.24$\times$)\\
& ogbn-arxiv & 71.68 $\pm$ 0.54 & \textit{54.64 $\pm$ 5.62 (5.04$\times$)}  & 71.80 $\pm$ 0.47 (5.09$\times$) & 71.47 $\pm$ 0.50 (6.14$\times$) & 71.21 $\pm$ 0.68 (6.98$\times$)\\
\multirow{4}{3em}{GCNII}
& Flickr & 52.37 $\pm$ 0.16     & 52.28 $\pm$ 0.16 (4.84$\times$)           & 52.31 $\pm$ 0.16 (4.91$\times$) & 52.36 $\pm$ 0.16 (5.54$\times$) & 52.23 $\pm$ 0.15 (6.44$\times$)\\
& Reddit & 96.32 $\pm$ 0.24     & \textit{86.50 $\pm$ 1.08 (4.51$\times$)}  & 96.11 $\pm$ 0.22 (4.52$\times$) & 96.01 $\pm$ 0.33 (5.16$\times$) & 95.54 $\pm$ 0.29 (5.92$\times$)\\
& Yelp   & 62.33 $\pm$ 0.20     & 62.21 $\pm$ 0.22 (5.26$\times$)           & 62.28 $\pm$ 0.26 (5.34$\times$) & 62.53 $\pm$ 0.36 (6.29$\times$) & 62.33 $\pm$ 0.37 (7.28$\times$)\\
& ogbn-arxiv & 72.52 $\pm$ 0.12 & \textit{44.57 $\pm$ 5.01 (6.54$\times$)}  & 72.28 $\pm$ 0.35 (6.74$\times$) & 72.22 $\pm$ 0.28 (7.98$\times$) & 71.74 $\pm$ 0.26 (10.24$\times$)\\
\hline
\multirow{4}{3em}{Bert-large} 
& MNLI  & 86.74 $\pm$ 0.24 & 85.98 $\pm$ 0.16 (7.55$\times$) & 86.61 $\pm$ 0.11 (7.38$\times$) & 86.68 $\pm$ 0.08 (9.13$\times$) & \textit{ 84.24 $\pm$ 0.74 (12.87$\times$)}\\
& SST-2 & 93.69 $\pm$ 0.30 & 93.46 $\pm$ 0.23 (7.55$\times$) & 93.54 $\pm$ 0.52 (7.30$\times$) & 93.20 $\pm$ 0.37 (9.05$\times$) & \textit{ 91.90 $\pm$ 1.04 (12.91$\times$)} \\
& MRPC  & 88.20 $\pm$ 0.02 & 87.36 $\pm$ 0.19 (7.55$\times$) & 87.90 $\pm$ 0.10 (7.40$\times$) & 87.69 $\pm$ 0.07 (9.19$\times$) & \textit{ 82.54 $\pm$ 0.38 (12.91$\times$)}\\
& QNLI  & 92.29 $\pm$ 0.14 & 92.34 $\pm$ 0.07 (7.55$\times$) & 92.44 $\pm$ 0.07 (7.42$\times$) & 92.43 $\pm$ 0.31 (9.19$\times$) & \textit{90.74 $\pm$ 0.13 (12.95$\times$)}\\
\bottomrule
\end{tabular}
}
\ifisarxiv
\else
\vspace{-1em}
\fi
\end{table*}

\subsection{Memory Saving and Computational Overhead}
\label{sec:mem-speed}

\begin{figure}[t]
	\centering
	\begingroup
	\setlength{\tabcolsep}{0pt} 
	\renewcommand{\arraystretch}{1} 
	\scriptsize
	\begin{tabular}{m{0.6cm}m{9cm}}
		\begin{minipage}{\linewidth}(a)\end{minipage} &\includegraphics[width=.85\linewidth]{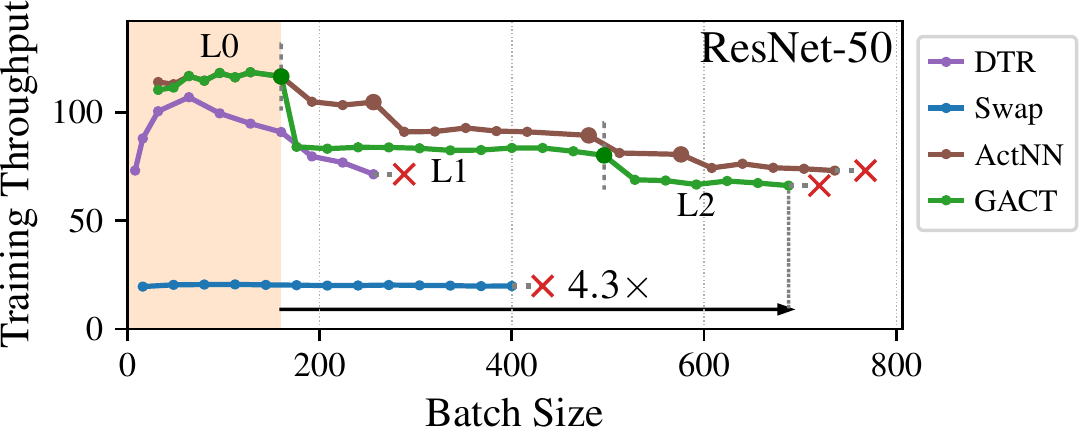}\\
		\begin{minipage}{\linewidth}(b)\end{minipage}
		&\includegraphics[width=.85\linewidth]{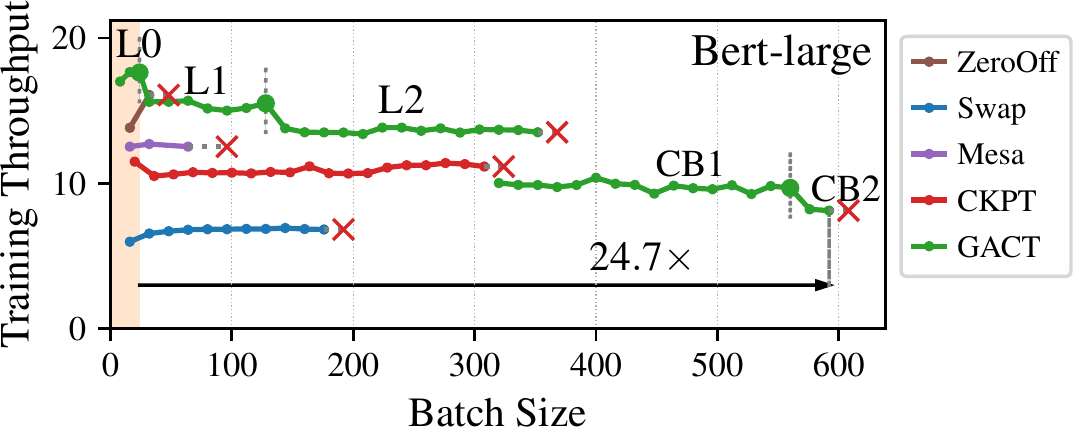}\\
		\begin{minipage}{\linewidth}(c)\end{minipage}
		&\includegraphics[width=.85\linewidth]{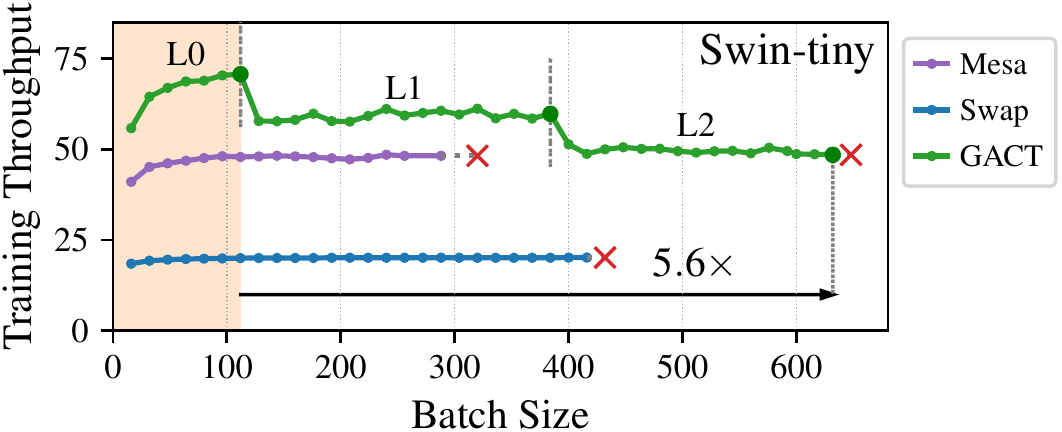}\\
	\end{tabular}
	\endgroup
\vspace{-1.5em}
\caption{Training throughput vs batch size. Red cross mark means out-of-memory. The shaded yellow region denotes the batch sizes with full precision training given the memory budget. CKPT: Gradient checkpointing, ZeroOff: ZeRO-Offload.
}
\label{fig:throughput_batch_size}
\end{figure}

\textbf{Settings and baselines.}
We implement the benchmark with PyTorch 1.10 and measure the memory saving and overhead of \method on an AWS g4dn.4xlarge instance, which has a 16GB NVIDIA T4 GPU and 64GB CPU memory.
On ResNet-50, we compare with ActNN~\cite{chen2021actnn}, a dedicated quantization framework for convolutional NNs, and DTR~\cite{kirisame2020dynamic}, a state-of-the-art rematerialization method for dynamic graphs. ``swap'' is a simple swapping strategy that swaps all activations to the CPU. 
For Bert-large, we also show the results on Mesa ~\cite{pan2021mesa}, a memory-saving resource-efficient training framework for transformers, and ZeRO-Offload~\cite{ren2021zero}, a highly optimized system for training large-scale language models. Gradient checkpointing uses the default checkpointing policy provided by the transformer library~\cite{wolf-etal-2020-transformers}, where only the input to each transformer block is saved before the backward pass.
On Swin-tiny, we only include Mesa and swap because other baselines lack the support for this network.

\textbf{Results.}
We compare the training throughput of \method against other memory saving systems in Fig.~\ref{fig:throughput_batch_size}.
On ResNet-50, \method achieves similar throughput as ActNN (ActNN optimization L5  is not listed because it optimizes PyTorch memory allocation, which is unrelated to quantization and can also be applied to \method), but ActNN enables training with a larger batch size. This is expected because ActNN implements efficient, customized layers for different operators in convolutional NNs.
For Bert-large, Zero-offload fails quickly because it only offloads optimizer states that occupy a small portion of total memory to CPU. \method L1 outperforms Mesa because Mesa only compresses tensors to 8 bit. When the batch is bigger, the activation size of each segment becomes the memory bottleneck and prevents gradient checkpointing from increasing the batch size. Moreover, combining \method with gradient checkpointing and efficient self-attention further reduces the peak memory, increasing the batch size by up to 24.7$\times$. Meanwhile, it introduces a small throughput overhead compared with the original gradient checkpointing. Across all the network architectures,  \method enables training with a 4.2$\times$ to 24.9$\times$ larger batch size under the same memory budget.

\textbf{Network scaling.}
With \method, we can construct larger models or train with a higher image resolution. Tab.~\ref{tab:scale} compares the largest model we can train against full precision. With the same batch size and memory budget, \method can scale a ResNet-152 to 7.0$\times$ deeper, 3.6$\times$ wider or 3.0$\times$ higher resolution. Similarly, Bert-large can be scaled to 2.0$\times$ deeper or 1.6$\times$ wider. In GCN, \method enables training 10.0$\times$ deeper and 1.7$\times$ wider network. Overall, \method maintains 75\% - 136\% original training throughput.

\begin{table}[t]
\caption{Largest models \method can train with 16G GPU memory. In ResNet (batch size=64), D (depth): number of layers, W (width): base width of the bottleneck block, R (resolution): width and height of input images. In Bert-large (batch size=16) and GCN, D (depth): number of transformer/gcn blocks, W (width): hidden size.}
\label{tab:scale}
\centering
\resizebox{\mywidth}{!}{%
\begin{tabular}{cc|ccc|ccc}
    \toprule
    &\multirow{2}{3em}{Dim} & \multicolumn{3}{|c|}{Maximum Value} 
                           & \multicolumn{3}{|c}{ Throughput (TFLOPS)}  \\
    & & FP & L1 & L2 & FP & L1 & L2 \\
    \midrule
    \multirow{3}{2em}{ResNet-152} 
    & D & 160 & 460 & 1124 & 0.43 & 0.47 & 0.41 \\
    & W & 88 & 304 & 320 & 0.44 & 0.89 & 0.6 \\
    & R & 232 & 548 & 716 & 0.41 & 0.39 & 0.44 \\
    \hline
    \multirow{2}{2em}{Bert-large} 
    & D & 32 & 56 & 64 & 0.67 & 0.56 & 0.53  \\
    & W & 1280 & 1488 & 2032 & 0.68 & 0.61 & 0.60  \\
    \hline
    \multirow{2}{2em}{GCN} 
    & D & 24 & 152 & 240 & 0.20 & 0.14 & 0.15 \\
    & W & 2464 & 3948 & 4244 & 0.36 & 0.38 & 0.40 \\
    \bottomrule
\end{tabular}%
}
\end{table}
\vspace{-1em}

\subsection{Other Optimizations}
\label{sec:swap}

\begin{table}[t]
\caption{Swap and prefetch speed/memory on Bert-large.}
\label{tab:swap}
\centering
\resizebox{\mywidth}{!}{%
\begin{tabular}{c|c|c|c}
    \toprule
    Algorithm & \shortstack{Speed\\(sequence/s)} & \shortstack{Peak Mem.\\ (MB)} & \shortstack{Total Mem.\\ (MB)} \\
    \midrule
     FP32                     & 16.41 & 9573 & 9527 \\
     FP32 + swap              & 6.02  & 5215 & 5093\\
     \method swap             & 12.95 & 5426 & 5325 \\
     \method swap + prefetch  & 14.02 & 5426 & 5324 \\
    \bottomrule
\end{tabular}%
}
\end{table}
We evaluate the idea of combining \method with swapping on Bert-large-cased. As shown in Tab.~\ref{tab:swap}, swapping compressed tensors is faster than swapping the original ones because communication between CPU and GPU is more time-consuming than computation. Combining \method with swapping increases training speed by up to 2.3$\times$. Notice here that the peak memory use of ``\method swap'' is slightly higher than ``FP32 + swap'' because \method does not quantize and swap intermediate states such as running mean/var of BatchNorm layer. Moreover, prefetch increases the speed by about 7\% with negligible memory overhead.

\begin{table}
\caption{Accuracy of Bert-large-cased on SST-2 and QNLI datasets}
\label{tab:acc-ckpt}
\centering
\resizebox{0.9\mywidth}{!}{%
\begin{tabular}{c c c|c c c}
    \toprule
    Algorithm & SST-2 & QNLI & Algorithm & SST-2 & QNLI\\
    \midrule
    FP32 & 93.58 & 92.42 &  CB1 & 93.81 & 92.26\\
    \bottomrule
\end{tabular}%
}
\end{table}

\begin{table}
\caption{Memory use of different algorithms on Bert-large. AM1: Activation size before backward, AM2: Activation size after reforwading the first transformer block. When batch size = 288, L0 runs out of memory, and therefore it is not listed below. }
\label{tab:mem-ckpt}
\centering
\resizebox{0.8\mywidth}{!}{%
\begin{tabular}{c|c|c|c|c}
    \toprule
    Batch Size & Algorithm & \shortstack{AM1\\(MB)} & \shortstack{AM2\\(MB)} & \shortstack{Peak Mem.\\(MB)} \\
    \midrule
    \multirow{4}{3em}{16} & L0 & 4434 & - & 9573 \\
    & FP32 + CKPT & 210 & 394 & 5541 \\
    & CB1 & 37 & 99 & 5286 \\
    & CB2 & 31 & 79 & 5269 \\
    \hline
    \multirow{4}{3em}{288} 
    & FP32 + CKPT & 3783 & 7092 & 12885 \\
    & CB1 & 515 & 1497 & 8251 \\
    & CB2 & 486 & 1307 & 8102 \\
    \bottomrule
\end{tabular}%
}
\end{table}

We next demonstrate combining \method with gradient checkpointing (CB1). Gradient checkpointing is performed at the beginning of each transformer block, thus avoiding saving tensors generated within the block. We then apply \method with gradient checkpointing, where the saved tensors are quantized with 4 bits. As shown in Tab.~\ref{tab:acc-ckpt}, the accuracy is unaffected. We also compare the activation memory and peak memory of CB1 and CB2 in Tab.~\ref{tab:mem-ckpt}. AM2 denotes the peak activation memory, which is the size of saved tensors after reforwarding the first transformer block. When batch size = 288, compared with gradient checkpointing on full precision (FP32), CB1 and CB2 reduce the peak activation size by 4.7$\times$ and 5.4$\times$ respectively.

%% file: aa-theorems.tex
\section{Proof of Theorems}

\subsection{Theorem 1: Convergence of ACT}
Assume that:\\
\noindent\textbf{A1.} $\Lc(\theta)$ is a continuous differentiable, $\nabla\Lc(\theta)$ is $\beta$-Lipschitz continuous.
.\\
\noindent\textbf{A2.} $\Lc(\theta)$ is bounded below by $\Lc_*$.\\
\noindent\textbf{A3.} $g(h; \theta)$ is differentiable w.r.t. $h$ and $\exists b>0$, s.t. $\forall \theta, \Eb\norm{g(Q(h(x, \theta)); \theta)-\hat g(h(x, \theta); \theta)}\le b$.\\
\noindent\textbf{A4.} $\exists \sigma^2>0$, s.t.,  $\forall\theta$, $\Var{\hat g(h(x, \theta)}\le \sigma^2$. \\
Then, for all $\eta < \frac{1}{2\beta}$, if we run ACT defined as Eq.~(\ref{eqn:ac}) for $T$ iterations, then we have 
\begin{align*}
\min_{t=0, \dots, T-1}\E{\norm{\nabla \Lc(\theta_t)}^2}
\le \frac{4(\Lc(\theta_0)-\Lc_*)}{\eta T}+3b^2+ \eta\beta\sigma^2
	\end{align*}

\begin{proof}
Denote $m:=\nabla_\theta \Lc(\theta_t)$, $\epsilon:=\hat g(h(x, \theta_t); \theta_t)-m$, $d:=g(Q(h(x;\theta_t));\theta_t)-\hat g(h(x, \theta_t); \theta_t)$. 
Then, by A3 and A4, we have
\begin{align}
\E{\epsilon}&=
\E{g(h(x, \theta_t); \theta_t) - \nabla_\theta \Lc(\theta_t)}+ 
\E{\iprod{J(x, \theta_t), 
\Delta Q(h(x, \theta_t))} }\nonumber\\
&=\iprod{J(x, \theta_t), 
\E{\Delta Q(h(x, \theta_t))}}=0. \label{eqn:epsilon}\\
\E{\norm{\epsilon}^2}&=\norm{\E{\epsilon}}^2
+\Var{\epsilon}=\Var{\hat g(h(x, \theta_t); \theta_t)}\le \sigma^2.\label{eqn:epsilon-sqr}\\
\E{\norm{d}}&\le b\label{eqn:d}.
\end{align}
By the definitions, the ACT dynamics can be written as
\begin{align*}
\theta_{t+1}\leftarrow \theta_t - \eta(m+d+\epsilon).
\end{align*}
By A1, we have
\begin{align}
\Lc(\theta_{t+1})\le \Lc(\theta_t) - \eta \iprod{m, m+d+\epsilon} + \frac{\beta\eta^2}{2}\norm{m+d+\epsilon}^2.\label{eqn:smoothness}
\end{align}
By Eq.s~(\ref{eqn:epsilon},\ref{eqn:d})
\begin{align}
\E{\iprod{m, m+d+\epsilon}}\ge\norm{m}^2 - \norm{m}\norm{d}+ \iprod{m, \E{\epsilon}}\ge\norm{m}^2-\norm{m}b
.\label{eqn:first-order-ineq}
\end{align}
By Eq.s~(\ref{eqn:epsilon},\ref{eqn:epsilon-sqr},\ref{eqn:d}), and $\norm{x+y}^2\le 2\norm{x}^2+2\norm{y}^2$,
\begin{align}
\E{\norm{m+d+\epsilon}^2}=\E{\norm{m+d}^2}+\Var{\epsilon}\le 2\E{\norm{m}}^2+2\E{\norm{d}}^2+\Var{\epsilon}=2\E{\norm{m}}^2+2b^2+\sigma^2.\label{eqn:second-order}
\end{align}
Taking expectation on both sides of Eq.~(\ref{eqn:smoothness}), plug in Eq.s~(\ref{eqn:first-order-ineq}, \ref{eqn:second-order}), and use $\eta<\frac{1}{2\beta}$, we have
\begin{align*}
\E{\Lc(\theta_{t+1})}\le& \Lc(\theta_t) - \eta (\norm{m}^2-\norm{m}b) + \frac{\beta\eta^2}{2}(2\E{\norm{m}}^2+2b^2+\sigma^2).\\
=&\Lc(\theta_t) - (\eta-\beta\eta^2)\norm{m}^2
+\eta \norm{m}b+\frac{\beta\eta^2}{2}(2b^2+\sigma^2)\\
=&\Lc(\theta_t) - \frac{\eta}{2}\norm{m}^2
+\eta \norm{m}b+\frac{\beta\eta^2}{2}(2b^2+\sigma^2).
\end{align*}
Completing the squares,
\begin{align*}
\E{\Lc(\theta_{t+1})}\le
\Lc(\theta_t) - \frac{\eta}{2}(\norm{m}-b)^2
+\frac{\beta\eta^2}{2}(2b^2+\sigma^2).
\end{align*}
Take expectation on both sides and sum up for $t={0, \dots, T-1}$, 
\begin{align*}
\E{\Lc(\theta_T)} - \Lc(\theta_0) \le -\frac{\eta}{2}\sum_{t=0}^{T-1}\Eb\left(\norm{\nabla \Lc(\theta_t)}-b\right)^2  +\frac{\beta\eta^2T}{2}(2b^2+\sigma^2).
\end{align*}
Reorganize the terms, 
\begin{align*}
\Es{t}{\Eb\left(\norm{\nabla \Lc(\theta_t)}-b\right)^2 }
\le \frac{2(\Lc(\theta_0)-\Lc(\theta_T))}{\eta T}+\eta\beta(2b^2+\sigma^2).
\end{align*}
Let $t_* = \argmin_t \E{\norm{\nabla \Lc(\theta_t)}}$, and use A1, we have
\begin{align*}
\Eb\left(\norm{\nabla\Lc(\theta_{t_*})}-b\right)^2\le 
\frac{2(\Lc(\theta_0)-\Lc_*)}{\eta T}+\eta\beta(2b^2+\sigma^2).
\end{align*}
Use $(a+b)^2\le 2a^2 + 2b^2$, we have
\begin{align*}
\E{\norm{\nabla\Lc(\theta_{t_*})}^2}\le 
\frac{4(\Lc(\theta_0)-\Lc_*)}{\eta T}+(2\beta\eta+2)b^2+ \eta\beta\sigma^2
\le \frac{4(\Lc(\theta_0)-\Lc_*)}{\eta T}+3b^2+ \eta\beta\sigma^2.
\end{align*}

\end{proof}

\subsection{Proposition 1: The Linearization Error}

\begin{proof}
Consider the gradient function $g(Q(h(x; \theta); \theta))$, whose output is a $P$-dimensional vector. 
Since it is twice differentiable, we construct the Taylor's expansion at $h(x; \theta)$ with Lagrange remainder:
\begin{align*}
\exists H_1, \dots, H_P, \mbox{s.t., }\forall i,~~
g_i(Q(h(x; \theta)); \theta) = 
g_i(h(x, \theta); \theta) + 
J_i(x, \theta)\Delta h(x, \theta)
+\Delta h(x, \theta)^\top H_i \Delta h(x, \theta),
\end{align*}
where $J_i(h(x; \theta), \theta):=\frac{\partial g_i(h(x; \theta); \theta)}{\partial h}$.
By the assumption, there exists $P>0$, such that the linearization error is
\begin{align*}
\norm{g(Q(h(x; \theta)); \theta) - \hat g(h(x; \theta); h(x; \theta), \theta)}_1
=\sum_{i=1}^P \Delta h(x, \theta)^\top H_i \Delta h(x, \theta)
\le \gamma  P \norm{\Delta h(x, \theta)}^2.
\end{align*}
Taking expectation,
\begin{align*}
&\E{\norm{g(Q(h(x; \theta)); h(x; \theta), \theta) - \hat g(h(x; \theta); \theta)}_2}\le 
\E{\norm{g(Q(h(x; \theta)); \theta) - \hat g(h(x; \theta); h(x; \theta), \theta)}_1}\\
&\le  \gamma  P \Var{\Delta h(x, \theta)}
=O(\Var{\Delta h(x, \theta)})
.
\end{align*}
\end{proof}

\subsection{Proposition 2: The Order of the Variance}
\newtheorem{innercustomthm}{Proposition}
\newenvironment{customthm}[1]
  {\renewcommand\theinnercustomthm{#1}\innercustomthm}
  {\endinnercustomthm}

The following proposition is convenient for isolating the different noise sources.
\begin{customthm}{A}
(Law of Total Variance)
$$\Var{X}=\E{\Varcond{X}{Y}} + \Var{\Econd{X}{Y}}.$$
\end{customthm}
\begin{proof}
By definition
\begin{align*}
\Var{\hat g(h(x; \theta_t); h(x; \theta), \theta_t)}
=\Var{g(h(x, \theta); \theta)} + 
\Var{J(h(x; \theta), \theta)\Delta h(x, \theta)},
\end{align*}
where $\Var{g(h(x, \theta); \theta)}$ is the noise introduced by subsampling the data $x$. By law of total variance, 
\begin{align*}
\Var{J(h(x; \theta), \theta)\Delta h(x, \theta)}= \Eb_{\Xc}\left[\Vars{Q}
{J(h(x; \theta); \theta_t)\Delta h(x, \theta)}
\right]+
\underbrace{
\Vars{\Xc}{\Es{Q}{J(h(x; \theta); \theta_t)\Delta h(x, \theta)}}}_{=0},
\end{align*}
where
\begin{align*}
\Vars{Q}{J(h(x; \theta); \theta_t)\Delta h(x, \theta)}
=&\Es{Q}{\norm{J(h(x; \theta); \theta_t)\Delta h(x, \theta)}^2}
\le \Es{Q}{\norm{J(h(x; \theta); \theta_t)}^2\norm{\Delta h(x, \theta)}^2}
\\=&
\norm{J(h(x; \theta); \theta_t)}^2
\Es{Q}{\norm{\Delta h(x, \theta)}^2}
= O\left(\Var{\Delta h(x, \theta)}\right).
\end{align*}
\end{proof}

\subsection{Proposition 3: The Structure of the Variance}\label{sec:appendix-var-structure}
Before investigating the structure of $\Vars{Q}
{J(x; \theta_t)\Delta h(x, \theta)}$, let's do some recap: 
the parameter $\theta_t$ is a $P$-dimensional vector; the context difference $\Delta h(x, \theta)$ is a $D$-dimensional vector, and $J(x; \theta_t)$ is a $P\times D$ matrix. Recall that $\Delta h(x, \theta)$ is the concatenation of $L$-vectors, $\Delta h^{(l)}(x, \theta)$, and let $J^{(l)}(x, \theta):=\frac{\partial g}{\partial h^{(l)}}g\left((h^{(l)}(x; \theta))_{l=1}^L, \theta\right)$, which is a $P\times D_l$ matrix. Furthermore, let $h^{(l)}_j(x, \theta)$ be the $j$-th dimension, and $J^{(l)}_j(x, \theta)$ be its $j$-th column.

To proceed, we need to make the following assumptions to the compressor $Q(\cdot): \Rb^{D}\rightarrow \Rb^{D}$:\\
\noindent\textbf{B1: } The compressed result is element-wise uncorrelated. That is, for any $i\ne j$, $\Cov{Q(h)_i}{Q(h)_j}=0$.\\
\noindent\textbf{B2: } For compressing a vector $h$ to $b$ bits, the compression variance of each dimension can be written in the form $\Var{Q(h)_j}\le R_j(h)S(b)$, where $S(\cdot)$ is a known function.

Both assumptions can be achieved by a stochastic rounding~\cite{courbariaux2015binaryconnect} quantizer, where
\begin{align*}
Q(h)_j = \begin{cases}
T_{h, b}^{-1}\left(\ceil{T_{h, b}(h_j)}\right) & \mbox{w.p. } T_{h, b}(h_j)-\floor{T_{h, b}(h_j)} \\
T_{h, b}^{-1}\left(\floor{T_{h, b}(h_j)}\right) & \mbox{otherwise}\\
\end{cases},
\end{align*}
where $T_{h, b}(h_j)=(2^b-1)\frac{h_j - \min_j h}{\max_j h-\min_j h}$. Since each dimension is quantized independently, B1 is met. Moreover, 
\begin{align*}
\Var{Q(h)_j}\le \frac{1}{4} \left(\frac{\max_j h-\min_j h}{(h_j - \min_j h)}\right)^2 (2^b-1)^{-2}=R_j(h) S(b),
\end{align*}
where
\begin{align*}
R_j(h) = \frac{1}{4} \left(\frac{\max_j h-\min_j h}{(h_j - \min_j h)}\right)^2,~~~ S(b) = (2^b-1)^{-2}.
\end{align*}

\begin{proof}
By definition, 
\begin{align*}
J(h; \theta)\Delta h = 
\sum_{l=1}^L \sum_{j=1}^{D_l} J^{(l)}_j(h; \theta_t) \Delta h^{(l)}_j.
\end{align*}
Using Assumption B1, we have
\begin{align*}
\Vars{Q}{J(h; \theta)\Delta h} &= 
\Es{Q}{\norm{\sum_{l=1}^L \sum_{j=1}^{D_l} J^{(l)}_j(h; \theta_t) \Delta h^{(l)}_j}^2}\\
&=\sum_{l=1}^L \sum_{j=1}^{D_l} \Es{Q}{\norm{J^{(l)}_j(h; \theta_t) \Delta h^{(l)}_j}^2}.\\
&=\sum_{l=1}^L \sum_{j=1}^{D_l} \norm{J^{(l)}_j(h; \theta_t)}^2 \Vars{Q}{\Delta h^{(l)}_j}
\end{align*}
Using Assumption B2, we have 
\begin{align*}
\Vars{Q}{J(h; \theta)\Delta h}
&\le\sum_{l=1}^L \sum_{j=1}^{D_l} \norm{J^{(l)}_j(h; \theta_t)}^2 R_l(h) S(b_l)
=\sum_{l=1}^L c_l(h, \theta) S(b_l),
\end{align*}
where $c_l(\theta, h) :=R_l(h) \norm{ J^{(l)}(h; \theta_t)}^2_F$.
\end{proof}

%% file: experiment-setup.tex
\section{Experiment Setup}
\label{sec:setup}
\subsection{Node classification task on graphs} We conduct experiments on four node classification datasets with standard splits, including Flickr, Reddit, Yelp from GraphSAINT~\cite{zeng2019graphsaint}, and ogbn-arxiv from Open Graph Benchmark (OGB)~\cite{hu2020open}. The four datasets cover extensive downstream applications with different scales. 
We use accuracy as the evaluation metric for multi-class classification and micro-F1 for multi-label classification. We run ten seeds (0 to 9) and report the average accuracy across~runs.

We evaluate \method on three representative GNN models, including GCN~\cite{kipf2016semi}, GAT~\cite{velivckovic2017graph}, and GCNII~\cite{chen2020simple_gcnii} under the full-batch training setting. 
All three models are implemented by CogDL~\cite{cen2021cogdl}, a toolkit for graph neural networks.

\subsection{Text classification task} We select four largest datasets, MNLI, QQP, SST-2, and QNLI, from the GLUE benchmark~\cite{wang2018glue}. The four datasets cover different aspects of natural language understanding, including sentiment classification, natural language inference and paraphrase detection. We use the mainstream transformer implementation~\cite{wolf-etal-2020-transformers} to train Bert-large~\cite{devlin2019bert}. We run three
seeds (42, 43, 44) and report F1 for QQP, accuracy for the others.

%% file: experiment-baseline-accuracy.tex
\section{Training Accuracy of Baselines}
\label{sec:baseline-acc}
For all the baselines we compared in Sec.~\ref{sec:mem-speed}, only ActNN, Mesa, and ZeRO-Offload are lossy methods. All other methods are lossless and have the same training accuracy as FP32. For ResNet-50 on ImageNet, the training accuracy for FP32, \method, ActNN L2, and ActNN L3 are 77.3, 77.0, 77.4, and 76.9. For Bert-Large on SST-2, the accuracy for FP32, \method, Mesa, and ZeRO-Offload are 93.7, 93.5, 93.8, and 93.3.
For Swin-tiny on ImageNet, the training accuracy for FP32, \method, and Mesa are 81.2, 81.0, and 81.3 respectively.

%% file: main.bbl
\begin{thebibliography}{10}

\bibitem{devlin2018bert}
Jacob Devlin, Ming-Wei Chang, Kenton Lee, and Kristina Toutanova.
\newblock Bert: Pre-training of deep bidirectional transformers for language
  understanding.
\newblock {\em arXiv preprint arXiv:1810.04805}, 2018.

\bibitem{fedus2021switch}
William Fedus, Barret Zoph, and Noam Shazeer.
\newblock Switch transformers: Scaling to trillion parameter models with simple
  and efficient sparsity.
\newblock {\em arXiv preprint arXiv:2101.03961}, 2021.

\bibitem{chakrabarti2019backprop}
Ayan Chakrabarti and Benjamin Moseley.
\newblock Backprop with approximate activations for memory-efficient network
  training.
\newblock {\em arXiv preprint arXiv:1901.07988}, 2019.

\bibitem{fu2020don}
Fangcheng Fu, Yuzheng Hu, Yihan He, Jiawei Jiang, Yingxia Shao, Ce~Zhang, and
  Bin Cui.
\newblock Don’t waste your bits! squeeze activations and gradients for deep
  neural networks via tinyscript.
\newblock In {\em International Conference on Machine Learning}, pages
  3304--3314. PMLR, 2020.

\bibitem{chen2021actnn}
Jianfei Chen, Lianmin Zheng, Zhewei Yao, Dequan Wang, Ion Stoica, Michael~W
  Mahoney, and Joseph~E Gonzalez.
\newblock Actnn: Reducing training memory footprint via 2-bit activation
  compressed training.
\newblock In {\em International Conference on Machine Learning}, 2021.

\bibitem{evans2021ac}
R~David Evans and Tor Aamodt.
\newblock {AC-GC}: Lossy activation compression with guaranteed convergence.
\newblock {\em Advances in Neural Information Processing Systems}, 34, 2021.

\bibitem{pan2021mesa}
Zizheng Pan, Peng Chen, Haoyu He, Jing Liu, Jianfei Cai, and Bohan Zhuang.
\newblock Mesa: A memory-saving training framework for transformers.
\newblock {\em arXiv preprint arXiv:2111.11124}, 2021.

\bibitem{evans2020jpeg}
R~David Evans, Lufei Liu, and Tor~M Aamodt.
\newblock Jpeg-act: accelerating deep learning via transform-based lossy
  compression.
\newblock In {\em 2020 ACM/IEEE 47th Annual International Symposium on Computer
  Architecture (ISCA)}, pages 860--873. IEEE, 2020.

\bibitem{jin2021novel}
Sian Jin, Guanpeng Li, Shuaiwen~Leon Song, and Dingwen Tao.
\newblock A novel memory-efficient deep learning training framework via
  error-bounded lossy compression.
\newblock In {\em 26th ACM SIGPLAN Symposium on Principles and Practice of
  Parallel Programming}, pages 485--487, 2021.

\bibitem{anonymous2022exact}
Anonymous.
\newblock {EXACT}: Scalable graph neural networks training via extreme
  activation compression.
\newblock In {\em Submitted to The Tenth International Conference on Learning
  Representations}, 2022.
\newblock under review.

\bibitem{vaswani2017attention}
Ashish Vaswani, Noam Shazeer, Niki Parmar, Jakob Uszkoreit, Llion Jones,
  Aidan~N Gomez, {\L}ukasz Kaiser, and Illia Polosukhin.
\newblock Attention is all you need.
\newblock In {\em Advances in neural information processing systems}, pages
  5998--6008, 2017.

\bibitem{kipf2016semi}
Thomas~N Kipf and Max Welling.
\newblock Semi-supervised classification with graph convolutional networks.
\newblock {\em arXiv preprint arXiv:1609.02907}, 2016.

\bibitem{micikevicius2018mixed}
Paulius Micikevicius, Sharan Narang, Jonah Alben, Gregory Diamos, Erich Elsen,
  David Garcia, Boris Ginsburg, Michael Houston, Oleksii Kuchaiev, Ganesh
  Venkatesh, and Hao Wu.
\newblock Mixed precision training.
\newblock In {\em International Conference on Learning Representations}, 2018.

\bibitem{wu2018training}
Shuang Wu, Guoqi Li, Feng Chen, and Luping Shi.
\newblock Training and inference with integers in deep neural networks.
\newblock In {\em International Conference on Learning Representations}, 2018.

\bibitem{wang2018training}
Naigang Wang, Jungwook Choi, Daniel Brand, Chia-Yu Chen, and Kailash
  Gopalakrishnan.
\newblock Training deep neural networks with 8-bit floating point numbers.
\newblock In {\em Advances in Neural Information Processing Systems}, pages
  7675--7684, 2018.

\bibitem{banner2018scalable}
Ron Banner, Itay Hubara, Elad Hoffer, and Daniel Soudry.
\newblock Scalable methods for 8-bit training of neural networks.
\newblock In {\em Advances in Neural Information Processing Systems}, pages
  5145--5153, 2018.

\bibitem{chen2020statistical}
Jianfei Chen, Yu~Gai, Zhewei Yao, Michael~W Mahoney, and Joseph~E Gonzalez.
\newblock A statistical framework for low-bitwidth training of deep neural
  networks.
\newblock In {\em Advances in neural information processing systems}, 2020.

\bibitem{sun2020ultra}
Xiao Sun, Naigang Wang, Chia-Yu Chen, Jiamin Ni, Ankur Agrawal, Xiaodong Cui,
  Swagath Venkataramani, Kaoutar El~Maghraoui, Vijayalakshmi~Viji Srinivasan,
  and Kailash Gopalakrishnan.
\newblock Ultra-low precision 4-bit training of deep neural networks.
\newblock In {\em Advances in Neural Information Processing Systems},
  volume~33, 2020.

\bibitem{chen2016training}
Tianqi Chen, Bing Xu, Chiyuan Zhang, and Carlos Guestrin.
\newblock Training deep nets with sublinear memory cost.
\newblock {\em arXiv preprint arXiv:1604.06174}, 2016.

\bibitem{jain2019checkmate}
Paras Jain, Ajay Jain, Aniruddha Nrusimha, Amir Gholami, Pieter Abbeel, Kurt
  Keutzer, Ion Stoica, and Joseph~E Gonzalez.
\newblock Checkmate: Breaking the memory wall with optimal tensor
  rematerialization.
\newblock {\em arXiv preprint arXiv:1910.02653}, 2019.

\bibitem{kirisame2020dynamic}
Marisa Kirisame, Steven Lyubomirsky, Altan Haan, Jennifer Brennan, Mike He,
  Jared Roesch, Tianqi Chen, and Zachary Tatlock.
\newblock Dynamic tensor rematerialization.
\newblock {\em arXiv preprint arXiv:2006.09616}, 2020.

\bibitem{huang2020swapadvisor}
Chien-Chin Huang, Gu~Jin, and Jinyang Li.
\newblock Swapadvisor: Pushing deep learning beyond the gpu memory limit via
  smart swapping.
\newblock In {\em Twenty-Fifth International Conference on Architectural
  Support for Programming Languages and Operating Systems}, pages 1341--1355,
  2020.

\bibitem{wang2018superneurons}
Linnan Wang, Jinmian Ye, Yiyang Zhao, Wei Wu, Ang Li, Shuaiwen~Leon Song,
  Zenglin Xu, and Tim Kraska.
\newblock Superneurons: Dynamic {GPU} memory management for training deep
  neural networks.
\newblock In {\em 23rd ACM SIGPLAN symposium on principles and practice of
  parallel programming}, pages 41--53, 2018.

\bibitem{peng2020capuchin}
Xuan Peng, Xuanhua Shi, Hulin Dai, Hai Jin, Weiliang Ma, Qian Xiong, Fan Yang,
  and Xuehai Qian.
\newblock Capuchin: Tensor-based gpu memory management for deep learning.
\newblock In {\em Twenty-Fifth International Conference on Architectural
  Support for Programming Languages and Operating Systems}, pages 891--905,
  2020.

\bibitem{beaumont2021efficient}
Olivier Beaumont, Lionel Eyraud-Dubois, and Alena Shilova.
\newblock Efficient combination of rematerialization and offloading for
  training dnns.
\newblock {\em Advances in Neural Information Processing Systems}, 34, 2021.

\bibitem{bottou2010large}
L{\'e}on Bottou.
\newblock Large-scale machine learning with stochastic gradient descent.
\newblock In {\em Proceedings of COMPSTAT'2010}, pages 177--186. Springer,
  2010.

\bibitem{bottou2018optimization}
L{\'e}on Bottou, Frank~E Curtis, and Jorge Nocedal.
\newblock Optimization methods for large-scale machine learning.
\newblock {\em SIAM Review}, 60(2):223--311, 2018.

\bibitem{courbariaux2015binaryconnect}
Matthieu Courbariaux, Yoshua Bengio, and Jean-Pierre David.
\newblock Binaryconnect: Training deep neural networks with binary weights
  during propagations.
\newblock In {\em Advances in neural information processing systems}, pages
  3123--3131, 2015.

\bibitem{rabe2021self}
Markus~N Rabe and Charles Staats.
\newblock Self-attention does not need $o(n^2)$ memory.
\newblock {\em arXiv preprint arXiv:2112.05682}, 2021.

\bibitem{simonyan2014very}
K.~Simonyan and A.~Zisserman.
\newblock Very deep convolutional networks for large-scale image recognition.
\newblock In {\em International Conference on Learning Representations}, 2015.

\bibitem{he2016deep}
Kaiming He, Xiangyu Zhang, Shaoqing Ren, and Jian Sun.
\newblock Deep residual learning for image recognition.
\newblock In {\em IEEE conference on computer vision and pattern recognition},
  pages 770--778, 2016.

\bibitem{liu2021Swin}
Ze~Liu, Yutong Lin, Yue Cao, Han Hu, Yixuan Wei, Zheng Zhang, Stephen Lin, and
  Baining Guo.
\newblock Swin transformer: Hierarchical vision transformer using shifted
  windows.
\newblock {\em International Conference on Computer Vision (ICCV)}, 2021.

\bibitem{imagenet_cvpr09}
Jia Deng, Wei Dong, Richard Socher, Li-Jia Li, Kai Li, and Li~Fei-Fei.
\newblock Imagenet: A large-scale hierarchical image database.
\newblock In {\em IEEE conference on computer vision and pattern recognition},
  pages 248--255. Ieee, 2009.

\bibitem{lin2017focal}
Tsung-Yi Lin, Priya Goyal, Ross Girshick, Kaiming He, and Piotr Doll{\'a}r.
\newblock Focal loss for dense object detection.
\newblock In {\em International Conference on Computer Vision (ICCV)}, pages
  2980--2988, 2017.

\bibitem{ren2015faster}
Shaoqing Ren, Kaiming He, Ross Girshick, and Jian Sun.
\newblock Faster {R-CNN}: Towards real-time object detection with region
  proposal networks.
\newblock {\em Advances in neural information processing systems}, 28:91--99,
  2015.

\bibitem{lin2014microsoft}
Tsung-Yi Lin, Michael Maire, Serge Belongie, James Hays, Pietro Perona, Deva
  Ramanan, Piotr Doll{\'a}r, and C~Lawrence Zitnick.
\newblock Microsoft coco: Common objects in context.
\newblock In {\em European conference on computer vision}, pages 740--755.
  Springer, 2014.

\bibitem{ren2021zero}
Jie Ren, Samyam Rajbhandari, Reza~Yazdani Aminabadi, Olatunji Ruwase, Shuangyan
  Yang, Minjia Zhang, Dong Li, and Yuxiong He.
\newblock Zero-offload: Democratizing billion-scale model training.
\newblock {\em arXiv preprint arXiv:2101.06840}, 2021.

\bibitem{wolf-etal-2020-transformers}
Thomas Wolf, Lysandre Debut, Victor Sanh, Julien Chaumond, Clement Delangue,
  Anthony Moi, Pierric Cistac, Tim Rault, Rémi Louf, Morgan Funtowicz, Joe
  Davison, Sam Shleifer, Patrick von Platen, Clara Ma, Yacine Jernite, Julien
  Plu, Canwen Xu, Teven~Le Scao, Sylvain Gugger, Mariama Drame, Quentin Lhoest,
  and Alexander~M. Rush.
\newblock Transformers: State-of-the-art natural language processing.
\newblock In {\em 2020 Conference on Empirical Methods in Natural Language
  Processing: System Demonstrations}, pages 38--45, Online, October 2020.
  Association for Computational Linguistics.

\bibitem{zeng2019graphsaint}
Hanqing Zeng, Hongkuan Zhou, Ajitesh Srivastava, Rajgopal Kannan, and Viktor
  Prasanna.
\newblock Graphsaint: Graph sampling based inductive learning method.
\newblock {\em arXiv preprint arXiv:1907.04931}, 2019.

\bibitem{hu2020open}
Weihua Hu, Matthias Fey, Marinka Zitnik, Yuxiao Dong, Hongyu Ren, Bowen Liu,
  Michele Catasta, and Jure Leskovec.
\newblock Open graph benchmark: Datasets for machine learning on graphs.
\newblock {\em arXiv preprint arXiv:2005.00687}, 2020.

\bibitem{velivckovic2017graph}
Petar Veli{\v{c}}kovi{\'c}, Guillem Cucurull, Arantxa Casanova, Adriana Romero,
  Pietro Lio, and Yoshua Bengio.
\newblock Graph attention networks.
\newblock {\em arXiv preprint arXiv:1710.10903}, 2017.

\bibitem{chen2020simple_gcnii}
Ming Chen, Zhewei Wei, Zengfeng Huang, Bolin Ding, and Yaliang Li.
\newblock Simple and deep graph convolutional networks.
\newblock In {\em International Conference on Machine Learning}, pages
  1725--1735. PMLR, 2020.

\bibitem{cen2021cogdl}
Yukuo Cen, Zhenyu Hou, Yan Wang, Qibin Chen, Yizhen Luo, Xingcheng Yao, Aohan
  Zeng, Shiguang Guo, Peng Zhang, Guohao Dai, Yu~Wang, Chang Zhou, Hongxia
  Yang, and Jie Tang.
\newblock Cogdl: Toolkit for deep learning on graphs.
\newblock {\em arXiv preprint arXiv:2103.00959}, 2021.

\bibitem{wang2018glue}
Alex Wang, Amanpreet Singh, Julian Michael, Felix Hill, Omer Levy, and Samuel
  Bowman.
\newblock Glue: A multi-task benchmark and analysis platform for natural
  language understanding.
\newblock In {\em 2018 EMNLP Workshop BlackboxNLP: Analyzing and Interpreting
  Neural Networks for NLP}, pages 353--355, 2018.

\bibitem{devlin2019bert}
Jacob Devlin, Ming-Wei Chang, Kenton Lee, and Kristina Toutanova.
\newblock Bert: Pre-training of deep bidirectional transformers for language
  understanding.
\newblock In {\em 2019 Conference of the North American Chapter of the
  Association for Computational Linguistics: Human Language Technologies,
  Volume 1 (Long and Short Papers)}, pages 4171--4186, 2019.

\end{thebibliography}
